\documentclass[runningheads]{llncs}

\usepackage[T1]{fontenc}
\usepackage{graphicx}
\usepackage{multirow}
\usepackage[dvipsnames]{xcolor}
\usepackage{etoolbox}
\usepackage{amsmath}
\usepackage{latexsym,amssymb}
\usepackage{upgreek}
\usepackage{algorithm}
\usepackage{algorithmicx}
\usepackage[noend]{algpseudocode}
\usepackage{array}
\usepackage{adjustbox}
\usepackage{cite}
\usepackage{soul}
\usepackage{url}
\usepackage[utf8]{inputenc}
\usepackage[small]{caption}
\usepackage{booktabs}
\usepackage{empheq}
\usepackage{fancybox}
\usepackage{subcaption}
\usepackage[english]{babel}
\usepackage{xspace}
\usepackage{setspace}
\usepackage{stmaryrd}
\usepackage{enumitem}
\usepackage{ifthen}
\usepackage{verbatim}
\usepackage{algorithmicx}
\let\llncssubparagraph\subparagraph
\let\subparagraph\paragraph
\usepackage{titlesec}
\let\subparagraph\llncssubparagraph

\usepackage{amsthm}
\usepackage{hyperref}
\usepackage[capitalise]{cleveref}

\usepackage{tikz}
\usepackage{forest}

\usetikzlibrary{arrows,decorations.pathmorphing,decorations.footprints,fadings,calc,trees,mindmap,shadows,decorations.text,patterns,positioning,shapes,matrix,fit}
\usetikzlibrary{arrows,shadows,backgrounds}
\usetikzlibrary{arrows.meta}
\usetikzlibrary{positioning,fit}
\usetikzlibrary{automata}
\usetikzlibrary{matrix}
\usetikzlibrary{shapes.symbols,shapes.misc,shapes.arrows,shapes.geometric}
\usetikzlibrary{matrix,chains,scopes,decorations.pathmorphing}

\newboolean{nonanon}               % Anonymous submission
\setboolean{nonanon}{true}        %
\newboolean{mkcompact}             % Compact submission
\setboolean{mkcompact}{false}     %
\newboolean{squeezeit}             % Apply baselinestretch
\setboolean{squeezeit}{false}     %
\crefname{prop}{Proposition}{Propositions}
\Crefname{prop}{Proposition}{Propositions}
\crefname{defn}{Definition}{Definitions}
\Crefname{defn}{Definition}{Definitions}
\crefname{cor}{Corollary}{Corollaries}
\Crefname{cor}{Corollary}{Corollaries}
\crefname{exmpl}{Example}{Examples}
\Crefname{exmpl}{Example}{Examples}

\newtheoremstyle{nutstyle}
{3.0pt} % Space above
{3.0pt} % Space below
{} % Body font
{} % Indent amount
{\bfseries} % Theorem head font
{.} % Punctuation after theorem head
{.5em} % Space after theorem head
{} % Theorem head spec (can be left empty, meaning `normal')

\theoremstyle{nutstyle}
\newtheorem{prop}{Proposition}
\newtheorem{defn}[prop]{Definition}
\newtheorem{cor}[prop]{Corollary}

\newtheoremstyle{nuxstyle}
{3.0pt} % Space above
{3.0pt} % Space below
{} % Body font
{} % Indent amount
{\itshape} % Theorem head font
{.} % Punctuation after theorem head
{.5em} % Space after theorem head
{} % Theorem head spec (can be left empty, meaning `normal')

\theoremstyle{nuxstyle}

\newcommand{\todoF}[2]{}

\newcommand{\fml}[1]{{\mathcal{#1}}}

\newcommand{\tn}[1]{\textnormal{#1}}
\newcommand{\mbf}[1]{\ensuremath\mathbf{#1}}
\newcommand{\mbb}[1]{\ensuremath\mathbb{#1}}

\newcommand{\mfrak}[1]{\ensuremath\mathfrak{#1}}

\newcommand{\tbf}[1]{\textbf{#1}}

\newcommand{\stwop}{\Upsigma_2^\tn{P}}

\newcommand{\SAT}{\ensuremath\mathsf{SAT}}
\newcommand{\outc}{\ensuremath\mathsf{outc}}
\newcommand{\True}{\textbf{true}}
\newcommand{\False}{\textbf{false}}

\newcommand{\prtwaxp}{\ensuremath\mathsf{reportWeakAXp}}
\newcommand{\newncl}{\ensuremath\mathsf{newNegCl}}
\newcommand{\newpcl}{\ensuremath\mathsf{newPosCl}}

\definecolor{darkslategray}{rgb}{0.18, 0.31, 0.31} %#2F4F4F
\definecolor{platinum}{rgb}{0.9, 0.89, 0.89} %#E5E4E2
\definecolor{gray}{rgb}{.4,.4,.4}
\definecolor{midgrey}{rgb}{0.5,0.5,0.5}
\definecolor{middarkgrey}{rgb}{0.35,0.35,0.35}
\definecolor{darkgrey}{rgb}{0.3,0.3,0.3}
\definecolor{darkred}{rgb}{0.7,0.1,0.1}
\definecolor{midblue}{rgb}{0.2,0.2,0.7}
\definecolor{darkblue}{rgb}{0.1,0.1,0.5}
\definecolor{defseagreen}{cmyk}{0.69,0,0.50,0}

\newcommand{\jnoteF}[1]{}

\newcommand{\oper}[1]{\ensuremath\textnormal{\small{\textsf{#1}}}}

\newcommand{\lval}{\oper{Lit}}
\newcommand{\lfeat}{\oper{Feat}}
\newcommand{\loper}{\oper{Oper}}
\newcommand{\childn}{\oper{children}}
\newcommand{\satisfy}{\oper{Sat}}
\newcommand{\leaf}{\oper{Leaf}}
\newcommand{\nleaf}{\oper{NonLeaf}}
\newcounter{tableeqn}[table]
\renewcommand{\thetableeqn}{\thetable.\arabic{tableeqn}}

\DeclareMathOperator*{\lequiv}{\leftrightarrow}
\DeclareMathOperator*{\limply}{\rightarrow}

\newcommand{\biglor}{\ensuremath\bigvee}
\newcommand{\bigland}{\ensuremath\bigwedge}
\newcommand{\waxp}{\ensuremath\mathsf{WAXp}}

\newcommand{\axp}{\ensuremath\mathsf{AXp}}

\newcommand{\mailtodomain}[1]{\href{mailto:#1@ciencias.ulisboa.pt}{\texttt{\nolinkurl{#1}}}}

\titleformat{\paragraph}[runin]
{\normalfont\bfseries}{}{0pt}{}

\newcolumntype{L}[1]{>{\raggedright\let\newline\\\arraybackslash\hspace{0pt}}m{#1}}
\newcolumntype{C}[1]{>{\centering\let\newline\\\arraybackslash\hspace{0pt}}m{#1}}
\newcolumntype{R}[1]{>{\raggedleft\let\newline\\\arraybackslash\hspace{0pt}}m{#1}}

\ifthenelse{\boolean{mkcompact}}{
\setlist{nolistsep}
\titlespacing*{\paragraph}{0pt}{3pt}{0.5em}[]

\setlength{\floatsep}{8pt plus 1.0pt minus 2.0pt}
\setlength{\textfloatsep}{10pt plus 1.0pt minus 2.0pt}
\setlength{\tabcolsep}{0.275em}

\setlength{\parskip}{0pt}

\setlength{\topsep}{0pt}
\makeatletter
\def\thm@space@setup{%
  \thm@preskip=0pt \thm@postskip=0pt
}
\makeatother

}{}

\ifthenelse{\boolean{squeezeit}}{

}{
  
}
{}
\AtBeginDocument{%
  \setlength{\floatsep}{8pt plus 1.0pt minus 2.0pt}
  \setlength{\textfloatsep}{8pt plus 1.0pt minus 2.0pt}
  \setlength{\belowdisplayskip}{3pt} \setlength{\belowdisplayshortskip}{3pt}
  \setlength{\abovedisplayskip}{3pt} \setlength{\abovedisplayshortskip}{3pt}
}

\makeatletter
\renewenvironment{proof}[1][\proofname]{\par
  \pushQED{\qed}%
  \normalfont \topsep0\p@\relax
  \trivlist
  \item[\hskip\labelsep\itshape#1\@addpunct{.}]\ignorespaces}{%
  \popQED\endtrivlist\@endpefalse
}

\newenvironment{Proof}[1][\proofname]{\par
  \pushQED{\qed}%
  \normalfont \partopsep=\z@skip \topsep=\z@skip
  \trivlist
  \item[\hskip\labelsep
        \itshape
    #1\@addpunct{.}]\ignorespaces
}{%
  \popQED\endtrivlist\@endpefalse
}
\makeatother

\begin{document}

\title{Feature Necessity \& Relevancy\\
  in ML Classifier Explanations %-- \\
  %Complexity \& Efficient Implementations
  %\thanks{Supported by organization x.}
}
\ifthenelse{\boolean{nonanon}}{
  \titlerunning{Feature Relevancy \& Necessity}
}{
  \titlerunning{}
}
%\titlerunning{Abbreviated paper title}
% If the paper title is too long for the running head, you can set
% an abbreviated paper title here
%

\ifthenelse{\boolean{nonanon}}{
  % ToDo: Authors & Affiliations
  % Order: Xuanxiang, Martin, Antonio, Jordi, Joao
  %
  \author{%
    Xuanxiang Huang\inst{1}\orcidID{0000-0002-3722-7191} \and
    Martin C.\ Cooper\inst{2}\orcidID{0000-0003-4853-053X} \and
    Antonio Morgado\inst{3}\orcidID{0000-0002-5295-1321} \and
    Jordi Planes\inst{4}\orcidID{0000-0003-1861-9736} \and
    Joao Marques-Silva\inst{5}\orcidID{0000-0002-6632-3086}
  }
  \authorrunning{X. Huang et al.}
  % First names are abbreviated in the running head.
  % If there are more than two authors, 'et al.' is used.
  %
  \institute{%
    University of Toulouse, France
    \email{xuanxiang.huang@univ-toulouse.fr}
    \and
    Univ.\ Paul Sabatier, IRIT, France
    \email{martin.cooper@irit.fr}
    \and
    Universitat de Lleida, Lleida, Spain
    \email{antonio.morgado@udl.cat}
    \and
    Universitat de Lleida, Lleida, Spain
    \email{jordi.planes@udl.cat}
    \and
    IRIT, CNRS, France
    \email{joao.marques-silva@irit.fr}
  }
}{
  \author{\tbf{Paper \# 094}}
  \authorrunning{~~}
  \institute{}
}

\maketitle              % typeset the header of the contribution

\begin{abstract}
  Given a machine learning (ML) model and a prediction, explanations
  can be defined as sets of features which are sufficient for the
  prediction.
  In some applications,
  and besides asking for an explanation, it is
  also critical to understand whether sensitive features can occur in
  some explanation, or whether a non-interesting feature must occur in
  all explanations.
  This paper starts by relating such queries respectively with the
  problems of relevancy and necessity in logic-based abduction.
  The paper then proves membership and hardness results for several
  families of ML classifiers.
  Afterwards the paper proposes concrete algorithms for two classes of
  classifiers. The experimental results confirm the scalability of the
  proposed algorithms.
  %\keywords{First keyword  \and Second keyword \and Another keyword.}
\end{abstract}

\section{Introduction} \label{sec:intro}

The remarkable achievements in machine learning (ML) in recent
years~\cite{bengio-nature15,bengio-bk16,bengio-cacm21} are not matched
by a comparable degree of trust. The most promising ML models are
inscrutable in their operation. As a direct consequence, the opacity
of ML models raises distrust in their use and deployment.
%%instead of helping to build trust. 
%
Motivated by a critical need for helping human decision makers to
grasp the decisions made by ML models, there has been extensive work
on explainable AI (XAI). Well-known examples include so-called model
agnostic explainers or alternatives based on saliency maps for neural
networks~\cite{muller-plosone15,guestrin-kdd16,lundberg-nips17,guestrin-aaai18}.
While most XAI approaches do not offer guarantees of rigor, and so can
produce explanations that are unsound given the underlying ML model,
there have been efforts on developing rigorous XAI approaches over the
last few years~\cite{darwiche-ijcai18,inms-aaai19,msi-aaai22}.
Rigorous explainability involves the computation of
explanations, but also the ability to answer a wide range of related
queries~\cite{marquis-kr20,marquis-kr21,hiims-kr21}.

By building on the relationship between explainability and logic-based 
abduction~\cite{gottlob-ese90,selman-aaai90,gottlob-jacm95,inms-aaai19},
this paper analyzes two concrete queries, namely feature necessity and
relevancy. Given an ML classifier, an instance (i.e.\ point in
feature space and associated prediction) and a target feature, the
goal of feature necessity is to decide whether the target feature
occurs in \emph{all} explanations of the given instance. Under the
same assumptions, the goal of feature relevancy is to decide whether a
feature occurs in \emph{some} explanation of the given instance.
This paper proves a number of complexity results regarding feature
necessity and relevancy, focusing on well-known families of
classifiers, some of which are widely used in ML. Moreover, the paper
proposes novel algorithms for deciding relevancy for two families of
classifiers.
The experimental results demonstrate the scalability of the proposed
algorithms.

The paper is organized as follows. The notation and definitions used
throughout are presented in~\cref{sec:prelim}. The problems of feature
necessity and relevancy are studied in~\cref{sec:frnt}, and example
algorithms are proposed in~\cref{sec:frna}.
\cref{sec:res} presents experimental results for a sample of families
of classifiers, \cref{sec:relw} relates our contribution with earlier work
and \cref{sec:conc} concludes the paper.

\section{Preliminaries} \label{sec:prelim}

\paragraph{Complexity classes, propositional logic \& quantification.}
The paper assumes basic knowledge of computational complexity, namely
the classes of decision problems P, NP and $\stwop$~\cite{arora-bk09}.
The paper also assumes basic knowledge of propositional logic,
including the Boolean satisfiability (SAT) problem for propositional
logic formulas in conjunctive normal form (CNF), 
and the use of SAT solvers as
oracles for the complexity class NP.
The interested reader is referred to textbooks on these
topics~\cite{arora-bk09,sat-hbk21}.

\subsection{Classification Problems}
Throughout the paper, we will consider classifiers as the underlying
ML model. 
Classification problems are defined on a set of features (or
attributes) $\fml{F}=\{1,\ldots,m\}$ and a set of classes
$\fml{K}=\{c_1,c_2,\ldots,c_K\}$.
Each feature $i\in\fml{F}$ takes values from a domain
$\mbb{D}_i$. Domains are categorical or ordinal, and each domain can be
defined on boolean, integer/discrete or real values. 
Feature space is defined as
$\mbb{F}=\mbb{D}_1\times{\mbb{D}_2}\times\ldots\times{\mbb{D}_m}$.
The notation $\mbf{x}=(x_1,\ldots,x_m)$ denotes an arbitrary point in
feature space, where each $x_i$ is a variable taking values from
$\mbb{D}_i$. The set of variables associated with the features is
$X=\{x_1,\ldots,x_m\}$.
Also the notation $\mbf{v}=(v_1,\ldots,v_m)$ represents a
specific point in feature space, where each $v_i$ is a constant
representing one concrete value from $\mbb{D}_i$.
A classifier $\mbb{C}$ is characterized by a (non-constant)
\emph{classification function} $\kappa$ that maps feature space
$\mbb{F}$ into the set of classes $\fml{K}$,
i.e.\ $\kappa:\mbb{F}\to\fml{K}$.
An \emph{instance}
denotes a pair $(\mbf{v}, c)$, where $\mbf{v}\in\mbb{F}$ and
$c\in\fml{K}$, with $c=\kappa(\mbf{v})$. 

\subsection{Examples of Classifiers}
The results presented in the paper apply to a comprehensive range of
widely used classifiers~\cite{flach-bk12,shalev-shwartz-bk14}. These
include,
decision trees (DTs)~\cite{breiman-bk84,iims-jair22},
decision graphs (DGs)~\cite{kohavi1994bottom} and diagrams (DDs)~\cite{akers1978binary,wegener-bk00},
decision lists (DLs)~\cite{rivest-ml87,ims-sat21} and sets (DSs)~\cite{clark1991rule,ipnms-ijcar18},
tree ensembles (TEs)~\cite{iisms-aaai22}, including random forests (RFs)~\cite{breiman2001random,ims-ijcai21}
and boosted trees (BTs)~\cite{friedman2001greedy},
neural networks (NNs)~\cite{muller1995neural}, naive bayes classifiers (NBCs)~\cite{kohavi1996scaling,msgcin-nips20},
classifiers represented with propositional languages, 
including deterministic decomposable negation normal form (d-DNNFs)~\cite{darwiche-jair02,hiicams-aaai22}
and its proper subsets, e.g. sentential decision diagrams (SDDs)~\cite{darwiche-ijcai11,broeck-aaai2015}
and free binary decision diagrams (FBDDs)~\cite{wegener-bk00, gergov1992efficient, darwiche-jair02},
and also monotonic classifiers.
In the rest of the paper, we will analyze some families of
classifiers in more detail.

\paragraph{d-DNNF classifiers.}
Negation normal form (NNF) is a well-known propositional language,
where the negation operators are restricted to atoms, or inputs. Any
propositional formula can de reduced to NNF in polynomial time.
Let the \emph{support} of a node be the set of atoms associated with
leaves reachable from the outgoing edges of the node.
Decomposable NNF (DNNF) is a restriction of NNF where the children of
AND nodes do not share atoms in their support.
A DNNF circuit is \emph{deterministic} (referred to as d-DNNF) if any
two children of OR nodes cannot both take value 1 for any assignment
to the inputs.
Restrictions of NNF including DNNF and d-DNNF exhibit important
tractability properties~\cite{darwiche-jair02}.
Besides, we briefly introduce FBDDs which is a proper subset of d-DNNFs.
An FBDD over a set $X$ of Boolean variables
is a rooted, directed acyclic graph comprising two types of nodes: \emph{nonterminal} and \emph{terminal}.
A nonterminal node is labeled by a variable $x_i \in X$, and has two outgoing edges, one labeled by 0 and the other by 1.
A terminal node is labeled by a 1 or 0, and has no outgoing edges.
For a subgraph rooted at a node labeled with a variable $x_i$, 
it represents a boolean function $f$ which is defined by the \emph{Shannon expansion}:
$f = (x_i \land f|_{x_i=1}) \lor (\neg x_i \land f|_{x_i=0})$,
where $f|_{x_i=1}$ ($f|_{x_i=0}$) denotes the \emph{cofactor}~\cite{brayton1984logic} of $f$ with respect to $x_i=1$ ($x_i=0$).
Moreover, any FBDD is \emph{read-once}, meaning that each variable is tested at most once on any path from the root node to a terminal node.

\paragraph{Monotonic classifiers.}
Monotonic classifiers find a number of important
applications, and have been studied extensively in recent
years~\cite{gupta-nips16,gupta-nips17,liu-nips20,vandenbroeck-nips20}.
Let $\preccurlyeq$ denote a partial order on the set of classes
$\fml{K}$. For example, we assume
$c_1\preccurlyeq{c_2}\preccurlyeq\ldots{c_K}$.
Furthermore, we assume that each domain $D_i$ is ordered such that the
value taken by feature $i$ is between a lower bound $\lambda(i)$ and
an upper bound $\mu(i)$. Given
$\mbf{v}_1=(v_{11},\ldots,v_{1i},\ldots,v_{1m})$ and
$\mbf{v}_2=(v_{21},\ldots,v_{2i},\ldots,v_{2m})$, we say that
$\mbf{v}_1\le\mbf{v}_2$ if
$\forall(i\in\fml{F}).(v_{1i}\le{v_{2i}})$.
Finally, a classifier is monotonic if whenever
$\mbf{v}_1\le\mbf{v}_2$, then
$\kappa(\mbf{v}_1)\preccurlyeq\kappa(\mbf{v}_2)$.

\paragraph{Running examples.}
As hinted above, throughout the paper, we will consider two fairly
different families of classifiers, namely classifiers represented with
d-DNNFs and monotonic classifiers.

\begin{figure}[t]
  %\begin{subfigure}[b]{1.0\textwidth}
  %  \begin{center}
  %    %\includegraphics[scale=0.125]{./figs/ddnnf01.jpg}
  %    \includegraphics[scale=0.125]{./figs/ddnnf02.jpg}
  %  \end{center}
  %  \caption{Graphical representation of d-DDNF}
  %\end{subfigure}
  %
  %\medskip
  %
  \begin{subfigure}[c]{0.5\textwidth}
    \begin{center}
      %\scalebox{0.9125}{\input{./texfigs/ddnnf02}}
      \scalebox{0.9125}{\begin{tikzpicture}[-,%
    node distance={1.25cm}, thin,
    nonleaf/.style = {draw, circle},
    leafn/.style = {draw, rectangle, minimum size=0.575cm},
    level 1/.style={sibling distance=35mm,level distance=1.125cm},
    level 2/.style={sibling distance=20mm,level distance=1.125cm},
    level 3/.style={sibling distance=15mm,level distance=1.125cm},
    ]
\node[nonleaf] (1) [label=above:{\small $n^k_1$}] {$\lor$}
child {
    node[nonleaf] (2) [label=above:{\small $n^k_2$}] {$\land$}
    child {
    node[leafn] (4) [label=above:{\small $n^k_4$}] {$x_1$}
    }
    child {
    node[nonleaf] (5) [label=above:{\small $n^k_5$}] {$\lor$}
    child {
    node[leafn] (8) [label=above:{\small $n^k_8$}] {$x_2$}
    }
    child {
    node[nonleaf] (9) [label=above:{\small $n^k_9$}] {$\land$}
    child {
    node[leafn] (12) [label=above:{\small $n^k_{12}$}] {$\neg{x_2}$}
    }
    child {
    node[leafn] (13) [label=above:{\small $n^k_{13}$}] {$x_4$}
    }
    }
    }
}
child {
    node[nonleaf] (3) [label=above:{\small $n^k_3$}] {$\land$}
    child {
    node[leafn] (6) [label=above:{\small $n^k_6$}] {$\neg{x_1}$}
    }
    child {
    node[nonleaf] (7) [label=above:{\small $n^k_7$}] {$\land$}
    child {
    node[leafn] (10) [label=above:{\small $n^k_{10}$}] {$x_3$}
    }
    child {
    node[leafn] (11) [label=above:{\small $n^k_{11}$}] {$x_4$}
    }
    }
}
;
\end{tikzpicture} }
    \end{center}
    \caption{Graphical representation of d-DDNF, i.e.\ $\kappa_1$}
  \end{subfigure}
  %
  %\medskip
  %
  \begin{minipage}[c]{0.5\textwidth}

    \bigskip%\medskip%
    
    \begin{subfigure}[c]{1.0\textwidth}
      \begin{center}
        \begin{tabular}{l}
          %\\[1.5pt]
          $\fml{F}_1=\{1,2,3,4\}$ \\[1.5pt]
          $\mbb{D}_{1i}=\{0,1\}, i=1,\ldots,4$\\[1.5pt]
          $\fml{K}_1=\{0,1\}$ \\
          %\\
        \end{tabular}
      \end{center}
      \caption{Definition of $\fml{F}_1,\mbb{D}_{1i},\fml{K}_1$}
    \end{subfigure}
    %

    %\hfill
    \bigskip%\medskip%
    %%
    %%\begin{subfigure}[c]{1.0\textwidth}
    %%  \quad
    %%\end{subfigure}
    %%
    \bigskip%\medskip%

    \begin{subfigure}[b]{1.0\textwidth}
      \begin{center}
      \scalebox{0.85}{
        \begin{tabular}{lccc}
          \tn{IF}      & $x_1=1 \land x_2=1$ & \tn{THEN} & 1 \\
          \tn{ELSE IF} & $x_1=1 \land x_4=1$ & \tn{THEN} & 1 \\
          \tn{ELSE IF} & $x_3=1 \land x_4=1$ & \tn{THEN} & 1 \\
          \tn{ELSE}    &         &           & 0 \\
        \end{tabular}
      }
      \end{center}
      \caption{Alternative representation of $\kappa_1$}
    \end{subfigure}
    %\begin{subfigure}[b]{1.0\textwidth}
    %  \begin{center}
    %    \begin{tabular}{lccc}
    %      \tn{IF}      & $x_1=1$ & \tn{THEN} & 1 \\
    %      \tn{ELSE IF} & $x_2=1$ & \tn{THEN} & 0 \\
    %      \tn{ELSE IF} & $x_3=1$ & \tn{THEN} & 0 \\
    %      \tn{ELSE IF} & $x_4=1$ & \tn{THEN} & 0 \\
    %      \tn{ELSE}    &         &           & 1 \\
    %    \end{tabular}
    %  \end{center}
    %  \caption{Alternative representation of $\kappa_1$}
    %\end{subfigure}
  \end{minipage} 
  \caption{Example of d-DDNF classifier} \label{fig:runex01}
\end{figure}
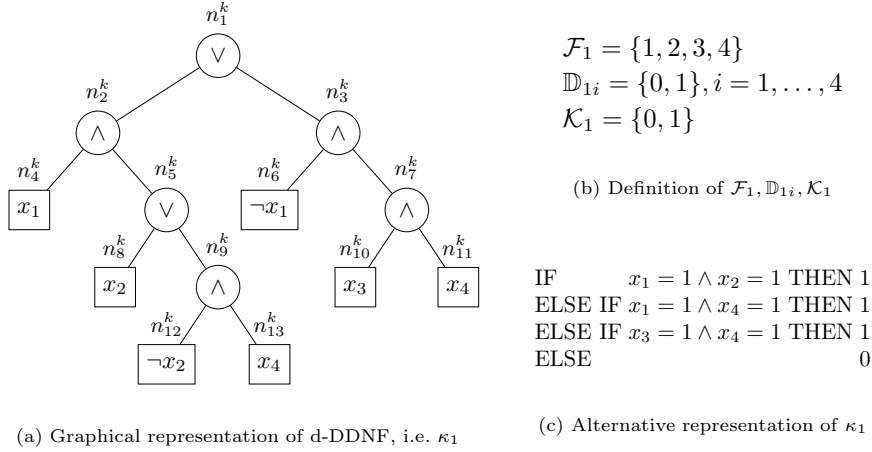

\begin{example} \label{ex:runex01}
  The first example is the d-DNNF classifier $\mbb{C}_1$ shown
  in~\cref{fig:runex01}.
  It represents the boolean function $(x_1 \land (x_2 \lor x_4)) \lor (\neg x_1 \land x_3 \land x_4)$.
  The instance considered throughout the paper is
  $(\mbf{v}_{1},c_{1})=((0,1,0,0),0)$.
\end{example}

\begin{figure}[t]
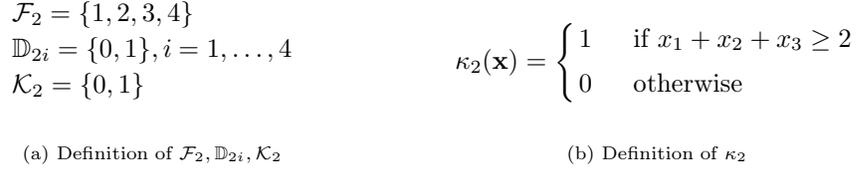

  \begin{subfigure}[b]{0.45\textwidth}
    \begin{center}
      \begin{tabular}{l}
        $\fml{F}_2=\{1,2,3,4\}$ \\[1.5pt]
        $\mbb{D}_{2i}=\{0,1\}, i=1,\ldots,4$\\[1.5pt]
        $\fml{K}_2=\{0,1\}$ \\
      \end{tabular}
    \end{center}
    \caption{Definition of $\fml{F}_2,\mbb{D}_{2i},\fml{K}_2$}
  \end{subfigure}
  \hfill
  \begin{subfigure}[b]{0.45\textwidth}
    \begin{center}
      \begin{tabular}{l}
        $\kappa_2(\mbf{x})=\left\{
        \begin{array}{ccl}
          1 & \quad\quad & \tn{if~$x_1+x_2+x_3\ge2$} \\[5pt] %\sum_{i=1}^{4}x_i
          0 & & \tn{otherwise}
        \end{array}\right.$\\
      \end{tabular}
    \end{center}
    \caption{Definition of $\kappa_2$}
  \end{subfigure}
  \caption{Example of a monotonic classifier} \label{fig:runex02}
\end{figure}

\begin{example} \label{ex:runex02}
  The second running example is the monotonic classifier $\mbb{C}_2$
%  Its classification function $\kappa_2(\mbf{x}) = (x_1+x_2+x_3\ge2$)
%  is defined on $\fml{F}_2=\{1,2,3,4\}$ (all features have same domain $\{0, 1\}$)
%  and $\fml{K}_2=\{0,1\}$.
  shown in~\cref{fig:runex02}.
  The instance that is considered throughout the paper is
  $(\mbf{v}_{2},c_{2})=((1,1,1,1),1)$.
\end{example}

\subsection{Formal Explainability} \label{ssec:fxai}
Prime implicant (PI) explanations~\cite{darwiche-ijcai18} represent a
minimal set of literals (relating a feature value $x_i$ and a constant 
$v_i\in\mbb{D}_i$) that are logically sufficient for the prediction.
PI-explanations are related with logic-based abduction, and so are
also referred to as abductive explanations (AXp's)~\cite{msi-aaai22}.
AXp's offer guarantees of rigor that are not offered by other
alternative explanation approaches.
More recently, AXp's have been studied
in terms of their computational
complexity~\cite{barcelo-nips20,marquis-kr21}.
There is a growing body of recent work on formal
explanations~\cite{darwiche-jair21,barcelo-nips21,kutyniok-jair21,kwiatkowska-ijcai21,mazure-cikm21,tan-nips21,rubin-aaai22,msi-aaai22,amgoud-ijcai22,leite-kr22,barcelo-corr22}.

Formally, given $\mbf{v}=(v_1,\ldots,v_m)\in\mbb{F}$, with
$\kappa(\mbf{v})=c$, an AXp is any subset-minimal set
$\fml{X}\subseteq\fml{F}$ such that,
\begin{equation} \label{eq:axp}
  \begin{array}{lcr}
    \waxp(\fml{X}) & \quad{~:=~}\quad\quad &
    \forall(\mbf{x}\in\mbb{F}).
    \left[
      \bigland\nolimits_{i\in{\fml{X}}}(x_i=v_i)
      \right]
    \limply(\kappa(\mbf{x})=c)
  \end{array}
\end{equation}
If a set $\fml{X}\subseteq\fml{F}$ is not %necessarily
minimal but \eqref{eq:axp} holds, then $\fml{X}$ is referred to as a
\emph{weak} AXp.
Clearly, the predicate $\waxp$ maps $2^{\fml{F}}$ into $\{\bot,\top\}$
(or $\{\False,\True\}$).
Given $\mbf{v}\in\mbb{F}$, an AXp $\fml{X}$ represents an irreducible
(or minimal) subset of the features which, if assigned the values
dictated by $\mbf{v}$, are sufficient for the prediction $c$,
i.e.\ value changes to the features not in $\fml{X}$ will not change
the prediction.
We can use the definition of the predicate $\waxp$ to formalize the
definition of the predicate $\axp$, also defined on subsets $\fml{X}$
of $\fml{F}$:
\begin{equation} \label{eq:axp2}
  \begin{array}{lcr}
    \axp(\fml{X}) & \quad{~:=~}\quad\quad &
    \waxp(\fml{X}) \land
    \forall(\fml{X}'\subsetneq\fml{X}).
    \neg\waxp(\fml{X}')
  \end{array}
\end{equation}
The definition of $\waxp(\fml{X})$ ensures that the predicate is 
\emph{monotone}.
Indeed, if $\fml{X}\subseteq\fml{X}'\subseteq\fml{F}$, and if
$\fml{X}$ is a weak AXp, then $\fml{X}'$ is also a weak AXp, as the
fixing of more features will not change the prediction.
Given the monotonicity of predicate $\waxp$, the definition of
predicate $\axp$ can be simplified as follows, with
$\fml{X}\subseteq\fml{F}$:
\begin{equation} \label{eq:axp3}
  \axp(\fml{X}) :=
  \waxp(\fml{X}) \land
  \forall(j\in\fml{X}).\neg\waxp(\fml{X}\setminus\{j\})
\end{equation}
This simpler but equivalent definition of AXp has important practical
significance, in that only a linear number of subsets needs to be
checked for, as opposed to exponentially many subsets
in~\eqref{eq:axp2}. As a result, the algorithms that compute one AXp
are based on~\eqref{eq:axp3}~\cite{msi-aaai22}.

\begin{example} \label{ex:runex01a}
  From~\cref{ex:runex01}, and given the instance $((0,1,0,0),0)$, we
  can conclude that the prediction will be 0 if features 1 and 3 take
  value 0, or if features 1 and 4 take value 0. Hence, the AXp's are
  $\{1,3\}$ and $\{1,4\}$. It is also apparent that the assignment
  $x_2=1$ bears no relevance on the fact that the prediction is 0.
\end{example}

\begin{example} \label{ex:runex02a}
  From~\cref{ex:runex02}, we can conclude that any sum of two
  variables assigned value 1 suffices for the prediction. Hence, given
  the instance $((1,1,1,1),1)$, the possible AXp's are $\{1,2\}$,
  $\{1,3\}$, and $\{2,3\}$. Observe that the definition of $\kappa_2$
  does not depend on feature 4.
\end{example}

Besides abductive explanations, another commonly studied type of
explanations are contrastive or counterfactual
explanations~\cite{miller-aij19,inams-aiia20,marquis-kr20,hiims-kr21}.
As argued in related work~\cite{hiims-kr21}, the duality between
abductive and contrastive explanations implies that for the purpose of
the queries studied in this paper, it suffices to study solely
abductive explanations.

\section{Feature Relevancy \& Necessity: Theory} \label{sec:frnt}

This section investigates the complexity of feature relevancy and
necessity\footnote{%
For the sake of brevity, we opt to only present sketches of some of the 
proofs.}.
We are interested in membership results, which allow us to devise
algorithms for the target problems. We are also interested in hardness
results, which serve to confirm that the running time complexities of
the proposed algorithms are within reason, given the problem's
complexity.

\subsection{Defining Necessity, Relevancy \& Irrelevancy}

Throughout this section, a classifier $\mbb{C}$ is assumed, with
features $\fml{F}$, domains $\mbb{D}_i$, $i\in\fml{F}$, classes
$\fml{K}$, a classification function $\kappa:\mbb{F}\to\fml{K}$,  and
a concrete instance $(\mbf{v},c)$, $\mbf{v}\in\mbb{F},c\in\fml{K}$. 

\begin{defn}[Feature Necessity, Relevancy \& Irrelevancy]
  Let $\mbb{A}$ denote the set of all AXp's for a classifier given
  a concrete instance, i.e.:
  \begin{equation}
  \mbb{A} = \{\fml{X}\subseteq\fml{F}\,|\,\axp(\fml{X})\}
  \end{equation}
  and let $t\in\fml{F}$ be a target feature.
  Then,
  \begin{enumerate}[nosep,label=\roman*.]
  \item $t$ is necessary if $t\in\cap_{\fml{X}\in\mbb{A}}\fml{X}$;
  \item $t$ is relevant if $t\in\cup_{\fml{X}\in\mbb{A}}\fml{X}$; and
  \item $t$ is irrelevant if
    $t\in\fml{F}\setminus\cup_{\fml{X}\in\mbb{A}}\fml{X}$.
  \end{enumerate}
\end{defn}
Throughout the remainder of the paper, the problem of deciding feature
necessity is represented by the acronym FNP, and the problem of deciding 
feature relevancy is represented by the acronym FRP.

\begin{example}
  As shown earlier, for the d-DNNF classifier of~\cref{fig:runex01},
  and given the instance $(\mbf{v}_1,c_1)=((0,1,0,0),0)$, there exist
  two AXp's, i.e.\  $\{1,3\}$ and $\{1,4\}$. Clearly, feature 1 is
  necessary, and features 1, 3 and 4 are relevant. In contrast,
  feature 2 is irrelevant.%\\
\end{example}

\begin{example}
  For the monotonic classifier of~\cref{fig:runex02}, and given the
  instance $(\mbf{v}_2,c_2)=((1,1,1,1),1)$, we have argued earlier
  that there exist three AXp's, i.e.\ $\{1,2\}$, $\{1,3\}$ and
  $\{2,3\}$, which allows us to conclude that features 1, 2 and 3 are
  relevant, but that feature 4 is irrelevant. In this case, there are
  no necessary features.
\end{example}

The general complexity of necessity and (ir)relevancy has been studied
in the context of logic-based
abduction~\cite{gottlob-ese90,selman-aaai90,gottlob-jacm95}.
Recent uses in explainability are briefly overviewed in~\cref{sec:relw}.

\subsection{Feature Necessity}

\begin{prop} \label{prop:fnp}
  If deciding $\waxp(\fml{X})$ is in complexity class $\mfrak{C}$, then
  FNP is in the complexity class co-$\mfrak{C}$.
\end{prop}

\begin{Proof}
  We first prove that there exists an AXp that does not include
  feature $t$ iff $\waxp(\fml{X})$ holds, with
  $\fml{X}=\fml{F}\setminus\{t\}$.
  \begin{itemize}[nosep]
  \item[$\Rightarrow$] Suppose there is some AXp $\fml{Z}$ that does
    not include $t$. Then, it must be the case that any superset of
    $\fml{Z}$ is a weak AXp.
    Thus, it must be true for the set
    $\fml{X}=\fml{F}\setminus\{t\}$. Hence, if such AXp $\fml{Z}$
    exists, then $\fml{X}$ must be a weak AXp.
  \item[$\Leftarrow$] Suppose that $\waxp(\fml{X})$ holds. Then there
    must exist at least one subset minimal set
    $\fml{Z}\subseteq\fml{X}$ that is an AXp, and such a set does not
    include $t$.
  \end{itemize}
  Thus we can decide feature necessity by a single call to 
  $\waxp(\fml{F}\setminus\{t\})$. 
  Since positive instances of FNP are negative instances of $\waxp(\fml{F}\setminus\{t\})$, 
  we conclude that FNP belongs to co-$\mfrak{C}$.
\end{Proof}

Given the known polynomial complexity of deciding
whether a set is a weak AXp for several families of
classifiers~\cite{msi-aaai22}, we then have the following result:

\begin{cor} \label{cor:P}
  For DTs, XpG's\footnote{%
  Explanation graphs (XpG's) have been proposed to enable the
  computation of explanations for decision graphs, and (multi-valued)
  decision diagrams~\cite{hiims-kr21}.}, NBCs, d-DNNF classifiers and
  monotonic classifiers, FNP is in P.
\end{cor}

\subsection{Feature Relevancy: Membership Results}

\begin{prop}[Feature Relevancy for DTs~\cite{hiims-kr21}]
  FRP for DTs is in P.
\end{prop}

\begin{prop} \label{prop:fmpnp}
  If deciding $\waxp(\fml{X})$ is in P, then FRP is in NP.
\end{prop}

\begin{Proof}
  Let $t\in\fml{F}$ be a target feature, and let
  $\fml{X}\subseteq\fml{F}$ be some guessed set of features, with
  $t\in\fml{X}$. To decide whether $\fml{X}$ is an AXp, we need to
  check that $\waxp(\fml{X})$ holds, which runs in polynomial time.
  Then, we must also check that, for all $i\in\fml{X}$,
  $\waxp(\fml{X}\setminus\{i\})$ does not hold, again in polynomial
  time. Hence, FRP is in NP.
\end{Proof}

The argument above can also be used for proving the following
results.

\begin{cor}
  For XpG's, NBCs, d-DNNF classifiers and monotonic classifiers, FRP is in NP.
\end{cor}

\begin{prop}
  If deciding $\waxp(\fml{X})$ is in NP, then FRP is in $\stwop$.
\end{prop}

\begin{cor}
  For DLs, DSs, RFs, BTs, and NNs, FRP is in $\stwop$. 
\end{cor}

\paragraph{Additional results.}
The following result will prove useful in designing algorithms for FRP
in practice.

\begin{prop} \label{prop:fmpnp2}
  Let $\fml{X}\subseteq\fml{F}$, and let $t\in\fml{X}$ denote some
  target feature such that, $\waxp(\fml{X})$ holds and
  $\waxp(\fml{X}\setminus\{t\})$ does not hold.
  Then, for any AXp $\fml{Z}\subseteq\fml{X}\subseteq\fml{F}$, it must
  be the case that $t\in\fml{Z}$.
\end{prop}

\begin{Proof}
  Let $\fml{Z}\subseteq\fml{F}$ be any AXp such that
  $\fml{Z}\subseteq\fml{X}$. Clearly, by definition $\waxp(\fml{Z})$
  must hold.
  Moreover, given the monotonicity of predicate $\waxp$,
  it is also the case that  $\waxp(\fml{Z}')$ must hold, with
  $\fml{Z}'=\fml{Z}\cup(\fml{X}\setminus(\fml{Z}\cup\{t\}))$, since
  $\fml{Z}\subseteq\fml{Z}'\subseteq\fml{F}$. However, by hypothesis,
  $\waxp(\fml{X}\setminus\{t\})$ does not hold; thus to avoid a contradiction 
  we must have $t \in \fml{Z}'$.
\end{Proof}

\subsection{Feature Relevancy: Hardness Results}

\begin{prop}[Relevancy for DNF Classifiers~\cite{hiims-kr21}]
  Feature relevancy for a DNF classifier is $\stwop$-hard.
\end{prop}

\begin{prop}%[Relevancy for Monotonic Classifiers]
  Feature relevancy for monotonic classifiers is NP-hard.
\end{prop}

\begin{Proof}
We say that a CNF is trivially satisfiable if some literal occurs in
all clauses. Clearly, SAT restricted to nontrivial CNFs is still
NP-complete. Let $\Phi$ be a not trivially satisfiable CNF on
variables $x_1,\ldots,x_k$. Let $N = 2k$. Let $\tilde{\Phi}$ be
identical to $\Phi$ except that each occurrence of a negative literal
$x_i$ ($1 \leq i \leq k$) is replaced by $x_{i+k}$. Thus
$\tilde{\Phi}$ is a CNF on $N$ variables each of which occur only
positively. Define the boolean classifier $\kappa$ (on $N+1$ features)
by $\kappa(x_0,x_1,\ldots,x_N) = 1$ iff $x_i = x_{i+k} = 1$ for some
$i \in \{1,\ldots,k\}$ or $x_0 \land \tilde{\Phi}(x_1,\ldots,x_N)=1$.
To show that $\Phi$ is monotonic we need to show that $\mbf{a} \leq
\mbf{b} \Rightarrow \kappa(\mbf{a}) \leq \kappa(\mbf{b})$. This 
follows by examining the two cases in which $\kappa(\mbf{a}) = 1$: if
$a_i=a_{i+k} \land \mbf{a} \leq \mbf{b}$, then $b_i=b_{i+k}$, whereas,
if $a_0 \land \tilde{\Phi}(a_1,\ldots,a_N)=1$ and $\mbf{a} \leq
\mbf{b}$,  then $b_0 \land \tilde{\Phi}(b_1,\ldots,b_N) = 1$ (by
positivity of $\tilde{\Phi}$), so in both cases $\kappa(\mbf{b}) = 1
\geq \kappa(\mbf{a})$.

Clearly $\kappa(\mbf{1}_{N+1}) = 1$. There are $k$ obvious AXp’s of
this prediction, namely $\{i, i+k\}$ ($1 \leq i \leq k$). These are
minimal by the assumption that $\Phi$ is not trivially
satisfiable. This means that no other AXp contains both $i$ and $i+k$
for any $i \in \{1,\ldots, k\}$. Suppose that $\Phi(\mbf{u})=1$. Let
$\fml{X}_u$ be $\{0\} \cup \{i \mid1 \leq i \leq k \land u_i=1\} \cup \{i+k
\mid 1 \leq i \leq k \land u_i=0\}$. Then $\fml{X}_u$ is a weak AXp of the
prediction $\kappa(1)=1$. Furthermore $\fml{X}_u$ does not contain any of
the AXp's $\{i,i+k\}$. Therefore some subset of $\fml{X}$ is an AXp and
clearly this subset must contain feature 0. Thus if $\Phi$ is
satisfiable, then there is an AXp which contains 0. 

We now show that the converse also holds. If $\fml{X}$ is an AXp of
$\kappa(\mbf{1}_{N+1}) = 1$ containing 0, then it cannot also contain
any of the pairs $i,i+k$ ($1 \leq i \leq k$), otherwise we could
delete 0 and still have an AXp.
We will show that this implies that we can build a satisfying
assignment $\mbf{u}$ for $\Phi$. Consider first
$\mbf{v}=(v_0,\ldots,v_N)$ defined by $v_i=1$ if $i \in \fml{X}$ ($0 \leq i
\leq N$) and $v_{i+k} = 1$ if neither $i$ nor $i+k$ belongs to $\fml{X}$ ($1
\leq i \leq k$), and $v_i=0$ otherwise ($1 \leq i \leq N$). Then
$\kappa(\mbf{v})=1$ by definition of an AXp, since $\mbf{v}$ agrees
with the vector $1$ on all features in $\fml{X}$. We can also note that
$v_0=1$ since $0 \in \fml{X}$. Since $\fml{X}$ does not contain $i$ and $i+k$ ($1
\leq i \leq k$), it follows that $v_i \neq v_{i+k}$. Now let $u_i=1$
iff $i \in \fml{X} \land 1 \leq i \leq k$. It is easy to verify that
$\Phi(\mbf{u}) = \tilde\Phi(\mbf{v}) = \kappa(\mbf{v}) = 1$. 

Thus, determining whether $\kappa(\mbf{1}_{N+1}) = 1$ has an AXp containing
the feature $0$ is equivalent to testing the satisfiability of
$\Phi$. It follows that FRP is NP-hard for monotonic classifiers by
this polynomial reduction from SAT.
\end{Proof}

\begin{prop} \label{prop:fmp-ddnnf}
Relevancy for FBDD classifiers is NP-hard.
\end{prop}

\begin{Proof}
Let $\psi$ be a CNF formula defined on a variable set $X = \{x_1, \dots, x_m\}$ and with clauses $\{\omega_1, \dots, \omega_n\}$.
We aim to construct an FBDD classifier $\fml{G}$ (representing a classification function $\kappa$) based on $\psi$ and a target variable in polynomial time,
such that: $\psi$ is SAT iff for $\kappa$ there is an AXp containing this target variable.

For any literal $l_j \in \omega_i$, replace $l_j$ with $l^i_j$.
Let $\psi' = \{\omega'_1, \dots, \omega'_n\}$ denote the resulting CNF formula
defined on the new variables $\{x^1_1, \dots, x^1_m, \dots x^n_1, \dots, x^n_m\}$.
For each original variable $x_j$,
let $I^+_j$ and $I^-_j$ denote the indices of clauses containing literal $x_j$ and $\neg x_j$, respectively.
So if $i \in I^+_j$, then $x^i_j \in \omega'_i$, if $i \in I^-_j$, then $\neg x^i_j \in \omega'_i$.
To build an FBDD $D$ from $\psi'$:
1) build an FBDD $D_i$ for each $\omega'_i$;
2) replace the terminal node 1 of $D_i$ with the root node of $D_{i+1}$;
$D$ is read-once because each variable $x^i_j$ occurs only once in $\psi'$.
Satisfying a literal $x^i_j \in \omega'_i$ means $x_j=1$,
while satisfying a literal $\neg x^k_j \in \omega'_k$ means $x_j=0$.
If both $x^i_j$ and $\neg x^k_j$ are satisfied, then it means we pick inconsistent values for the variable $x_j$, which is unacceptable.
Let us define $\phi$ to capture inconsistent values for any variable $x_j$:
\begin{align}
\phi := \biglor\nolimits_{1 \le j \le m} \left( \left(\biglor\nolimits_{i \in I^+_j} x^i_j \right) \land \left(\biglor\nolimits_{k \in I^-_j} \neg x^k_j \right) \right)
\end{align}
If $I^+_j = \emptyset$, then let $\left(\biglor\nolimits_{i \in I^+_j} x^i_j\right) = 0$.
If $I^-_j = \emptyset$, then let $\left(\biglor\nolimits_{k \in I^-_j} \neg x^k_j\right) = 0$.
Any true point of $\phi$ means we pick inconsistent values for some variable $x_j$,
so it represents an unacceptable point of $\psi$.
To avoid such inconsistency, one needs to at least falsify either
$\biglor\nolimits_{i \in I^+_j} x^i_j$ or $\biglor\nolimits_{k \in I^-_j} \neg x^k_j$ for each variable $x_j$.
To build an FBDD $G$ from $\phi$:
1) build FBDDs $G^+_j$ and $G^-_j$ for $\biglor\nolimits_{i \in I^+_j} x^i_j$ and $\biglor\nolimits_{k \in I^-_j} \neg x^k_j$, respectively;
2) replace the terminal node 1 of $G^+_j$ with the root node of $G^-_j$, let $G_j$ denote the resulting FBDD;
3) replace the terminal 0 of $G_j$ with the root node of $G_{j+1}$;
$G$ is read-once because each variable $x^i_j$ occurs only once in $\phi$.

Create a root node labeled $x^0_0$, link its 1-edge to the root of $D$, 0-edge to the root of $G$.
The resulting graph $\fml{G}$ is an FBDD representing $\kappa:= (x^0_0 \land \psi') \lor (\neg x^0_0 \land \phi)$,
$\kappa$ is a boolean classifier defined on $\{x^0_0, x^1_1, \dots, x^n_m\}$ and $x^0_0$ is the target variable.
The number of nodes of $\fml{G}$ is $O(n \times m)$.
Let $\fml{I} = \{(0,0), (1,1), \dots (n,m)\}$ denote the set of variable indices, for variable $x^i_j$, $(i,j) \in \fml{I}$.

Pick an instance $\mbf{v} = \{v^0_0, \dots, v^i_j, \dots\}$ satisfying every literal of $\psi'$ (i.e. $v^i_j=1$ and $v^k_j=0$ for $x^i_j, \neg x^k_j \in \psi'$)
and such that $v^0_0=1$, then $\psi' (\mbf{v}) = 1$, and so $\kappa(\mbf{v}) = 1$.
Suppose $\fml{X} \subseteq \fml{I}$ is an AXp of $\mbf{v}$:
1) If $\{(i, j), (k, j)\} \subseteq \fml{X}$ for some variable $x_j$, where $i \in I^+_j$ and $k \in I^-_j$,
then for any point $\mbf{u}$ of $\kappa$ such that $u^i_j = v^i_j$ for any $(i,j) \in \fml{X}$, we have $\kappa(\mbf{u}) = 1$ and $\phi(\mbf{u}) = 1$.
Moreover, if $\mbf{u}$ sets $u^0_0=1$, then $\kappa(\mbf{u}) = 1$ implies $\psi'(\mbf{u}) = 1$,
else if $\mbf{u}$ sets $u^0_0=0$, then $\kappa(\mbf{u}) = 1$ because of $\phi(\mbf{u}) = 1$.
$\kappa (\mbf{u}) = 1$ regardless the value of $u^0_0$, so $(0,0) \not \in \fml{X}$.
2) If $\{(i, j), (k, j)\} \not \subseteq \fml{X}$ for any variable $x_j$, where $i \in I^+_j$ and $k \in I^-_j$,
then for some point $\mbf{u}$ of $\kappa$ such that $u^i_j = v^i_j$ for any $(i,j) \in \fml{X}$,
we have $\phi (\mbf{u}) \neq 1$, in this case $\kappa(\mbf{u}) = 1$ implies $\psi'(\mbf{u}) = 1$,
besides, any such $\mbf{u}$ must set $u^0_0=1$, so $(0,0) \in \fml{X}$.

If case 2) occurs, then $\psi$ is satisfiable.
(a satisfying assignment is $x_j=1$ iff $\exists i \in I_j^{+}$ s.t. $(i,j) \in \fml{X}$).
If case 2) never occurs, then $\psi$ is unsatisfiable.
It follows that FRP is NP-hard for FBDD classifiers by this polynomial reduction from SAT.
\end{Proof}

\begin{cor}
Relevancy for d-DNNF classifiers is NP-hard.
\end{cor}

\section{Feature Relevancy: Example Algorithms} \label{sec:frna}
This section details two methods for FRP.
One method decides feature relevancy for d-DNNF classifiers,
whereas the other method decides feature relevancy for arbitrary monotonic classifiers.
Based on \cref{prop:fnp} and \cref{cor:P},
existing algorithm for computing one AXp~\cite{msgcin-nips20,msgcin-icml21,hiims-kr21,hiicams-aaai22}
can be used to decide feature necessity.
Hence, there is no need for devising new algorithms.
Additionally, the weak AXp returned from the proposed methods (if it exist)
can be fed (as a seed) into the algorithms of computing one AXp~\cite{msgcin-icml21,hiicams-aaai22}
to extract one AXp in polynomial time.

\subsection{Relevancy for d-DNNF Classifiers}

This section details a propositional encoding that decides feature relevancy for d-DNNFs.
The encoding follows the approach described in the proof of \cref{prop:fmpnp2},
and comprises two copies ($\mbb{C}^0$ and $\mbb{C}^t$) of the same d-DNNF classifier $\mbb{C}$,
$\mbb{C}^0$ encodes $\waxp(\fml{X})$ (i.e. the prediction of $\kappa$ remains unchanged),
$\mbb{C}^t$ encodes $\neg\waxp(\fml{X}\setminus\{t\})$ (i.e. the prediction of $\kappa$ changes).
The encoding is polynomial in the size of classifier's representation.

\begin{table}[t]
\caption{Encoding for deciding whether there is a weak AXp including feature $t$.}
\label{tab:enc_ddnnf}
\begin{center}
    \scalebox{0.95}{
    \renewcommand{\arraystretch}{1.275}
    \renewcommand{\tabcolsep}{0.5em}
    \begin{tabular}{c|c|c} \toprule
    Conditions & Constraints & Fml~\# \\
    \midrule
    $\leaf(j), \lfeat(j,i), \satisfy(\lval(j),v_i)$ &
    $n^k_j$
    & \refstepcounter{tableeqn} {\small(\thetableeqn)}\label{eq101}
    \\[1.25pt]
    \cline{1-3}
    $\leaf(j),\lfeat(j,i),\neg\satisfy(\lval(j),{v_i}), i = k$ &
    $n^k_j$
    & \refstepcounter{tableeqn} {\small(\thetableeqn)}\label{eq102}
    \\[1.25pt]
    \cline{1-3}
    $\leaf(j),\lfeat(j,i),\neg\satisfy(\lval(j),{v_i}), i \neq k$ &
    $n^k_j \lequiv \neg{s_i}$
    & \refstepcounter{tableeqn} {\small(\thetableeqn)}\label{eq103}
    \\[1.25pt]
    \cline{1-3}
    $\nleaf(j), \loper(j)=\lor$  &
    $n^k_j \lequiv \biglor_{l\in\childn(j)} n^k_l$
    & \refstepcounter{tableeqn} {\small(\thetableeqn)}\label{eq104}
    \\[1.25pt]
    \cline{1-3}
    $\nleaf(j), \loper(j)=\land$  &
    $n^k_j \lequiv \bigland_{l\in\childn(j)} n^k_l$
    & \refstepcounter{tableeqn} {\small(\thetableeqn)}\label{eq105}
    \\[1.25pt]
    \cline{1-3}
    $\kappa(\mbf{v})=0$ &
    $\neg n^0_1$
    & \refstepcounter{tableeqn} {\small(\thetableeqn)}\label{eq106}
    \\
    \cline{1-3}
    $\kappa(\mbf{v})=0$ &
    $s_i \lequiv n^i_1$
    & \refstepcounter{tableeqn} {\small(\thetableeqn)}\label{eq107}
    \\
    \cline{1-3}
    &
    $s_t$
    & \refstepcounter{tableeqn} {\small(\thetableeqn)}\label{eq108}
    \\
    \bottomrule
    \end{tabular}
    }
\end{center}
\end{table}

The encoding is applicable to the case $\kappa(\mbf{x}) = 0$.
The case $\kappa(\mbf{x}) = 1$ can be transformed to $\neg \kappa(\mbf{x}) = 0$,
so we assume both d-DNNF $\mbb{C}$ and its negation $\neg \mbb{C}$ are given.
To present the constraints included in this encoding,
we need to introduce some auxiliary boolean variables and predicates.
\begin{enumerate}[nosep]
\item $s_i$, $1 \le i \le m$. $s_i$ is a selector
    such that $s_i = 1$ iff feature $i$ is included in a weak AXp candidate $\fml{X}$.
\item $n^k_j$, $1 \le j \le |\mbb{C}|$ and $0 \le k \le m$. $n^k_j$ is 
    the indicator of a node $j$ of d-DNNF $\mbb{C}$ for replica $k$.
    The indicator for the root node of $k$-th replica is $n^k_1$.
    Moreover, the semantics of  $n_j^k$
    is $n_j^k = 1$ iff the sub-d-DNNF rooted at node $j$ in $k$-th replica
    is consistent. 
\item
    $\leaf(j) = 1$ if the node $j$ is a leaf node.
\item
    $\nleaf(j) = 1$ if the node $j$ is a non-leaf node.
\item
    $\lfeat(j,i) = 1$ if the leaf node $j$ is labeled with feature $i$.
\item
    $\satisfy(\lval(j),v_i) = 1$
    if for leaf node $j$, the literal on feature $i$ is satisfied by $v_i$.
\end{enumerate}
The encoding is summarized in \autoref{tab:enc_ddnnf}.
As literals are d-DNNF leafs, the values of the selector variables
only affect the values of the indicator variables of leaf nodes. 
Constraint \eqref{eq101} states that
for any leaf node $j$ whose literal is consistent with
the given instance, its indicator $n^k_j$ is always consistent
regardless of the value of $s_i$.
On the contrary,
constraint \eqref{eq103} states that
for any leaf node $j$ whose literal is inconsistent with
the given instance, its indicator $n^k_j$ is consistent iff feature $i$ is not picked,
in other words, feature $i$ can take any value.
Because replica $k$ ($k > 0$) is used to check the necessity of
including feature $k$ in $\fml{X}$,
we assume the value of the local copy of selector $s_k$ is 0 in replica $k$.
In this case, as defined in constraint \eqref{eq102}, 
even though leaf node $j$ labeled feature $k$ has a literal
that is inconsistent with the given instance, its indicator $n^k_j$ is
consistent.
Constraint \eqref{eq104} defines
the indicator for an arbitrary $\lor$ node $j$.
Constraint \eqref{eq105} defines
the indicator for an arbitrary $\land$ node $j$.
Together, these constraints declare how the consistency is propagated
through the entire d-DNNF.
Constraint \eqref{eq106} states that
the prediction of the d-DNNF classifier $\mbb{C}$ remains $0$
since the selected features form a weak AXp.
Constraint \eqref{eq107} states that
if feature $i$ is selected, then removing it
will change the prediction of $\mbb{C}$.
Finally, constraint \eqref{eq108} indicates that feature $t$ must be
included in $\fml{X}$.

\begin{example}
Given the d-DNNF classifier of~\cref{fig:runex01} and the instance $(\mbf{v}_1,c_1)=((0,1,0,0),0)$,
suppose that the target feature is $3$.
We have selectors $\mbf{s} = \{s_1, s_2, s_3, s_4\}$, and the encoding is as follows:
\begin{enumerate}[nosep]
\item $
	(n^0_1 \lequiv n^0_2 \lor n^0_3) \land
	(n^0_2 \lequiv n^0_4 \land n^0_5) \land
	(n^0_3 \lequiv n^0_6 \land n^0_7) \land
	(n^0_5 \lequiv n^0_8 \lor n^0_9) \land \\
	(n^0_7 \lequiv n^0_{10} \land n^0_{11}) \land
	(n^0_9 \lequiv n^0_{12} \land n^0_{13}) \land
	(n^0_4 \lequiv \neg s_1) \land
	(n^0_6 \lequiv 1) \land 
	(n^0_8 \lequiv 1) \land 
	(n^0_{10} \lequiv \neg s_3) \land
	(n^0_{11} \lequiv \neg s_4) \land
	(n^0_{12} \lequiv \neg s_2) \land
	(n^0_{13} \lequiv \neg s_4) \land
	(\neg n^0_1) \land (s_3)
	$
\item $
	(n^3_1 \lequiv n^3_2 \lor n^3_3) \land
	(n^3_2 \lequiv n^3_4 \land n^3_5) \land
	(n^3_3 \lequiv n^3_6 \land n^3_7) \land
	(n^3_5 \lequiv n^3_8 \lor n^3_9) \land \\
	(n^3_7 \lequiv n^3_{10} \land n^3_{11}) \land
	(n^3_9 \lequiv n^3_{12} \land n^3_{13}) \land
	(n^3_4 \lequiv \neg s_1) \land
	(n^3_6 \lequiv 1) \land 
	(n^3_8 \lequiv 1) \land 
	(n^3_{10} \lequiv 1) \land
	(n^3_{11} \lequiv \neg s_4) \land
	(n^3_{12} \lequiv \neg s_2) \land
	(n^3_{13} \lequiv \neg s_4) \land
	(s_3 \lequiv n^3_1)
	$
\end{enumerate}
Given the AXp's listed in~\cref{ex:runex01a}, by solving these
formulas we will either obtain $\{1, 3\}$ or $\{1,4\}$ as the AXp.
\end{example}

\subsection{Relevancy for Monotonic Classifiers}

This section describes an algorithm for FRP in
the case of monotonic classifiers. No assumption is made regarding the
actual implementation of the monotonic classifier.

\paragraph{Abstraction refinement for relevancy.}
The algorithm proposed in this section iteratively refines an
over-approximation (or abstraction) of all the subsets $\fml{S}$ of
$\fml{F}$ such that: i) $\fml{S}$ is a weak AXp, and ii) any AXp
included in $\fml{S}$ also includes the target feature $t$.
Formally, the set of subsets of $\fml{F}$ that we are interested in is
defined as follows:
\begin{equation}
  \mbb{H} = \{\fml{S}\subseteq\fml{F}\,|\,
  \waxp(\fml{S})\land
  \forall(\fml{X}\subseteq\fml{S}).%
  \left[\axp(\fml{X})\limply(t\in\fml{X})\right]%
  \}
\end{equation}
The proposed algorithm iteratively refines the over-approximation of
set $\mbb{H}$ until one can decide with certainty whether $t$ is
included in some AXp. The refinement step involves exploiting
counterexamples as these are identified.
(The approach is referred to as
abstraction refinement FRP,
since the use of 
abstraction refinement can be related with earlier work (with the same
name) in model checking%
~\cite{clarke-jacm03}.)
In practice, it will in general be impractical to manipulate such
over-approximation of set $\mbb{H}$ explicitly. As a result, we use a
propositional formula (in fact a CNF formula) $\fml{H}$, such that the
models of $\fml{H}$ encode the subsets of features about which we have
yet to decide whether each of those subsets only contains AXp's that
include $t$. (Formula $\fml{H}$ is defined on a set of Boolean
variables $\{s_1,\ldots,s_m\}$, where each $s_i$ is associated with
feature $i$, and assigning $s_i=1$ denotes that feature $i$ is
included in a given set, as described below.)
The algorithm then iteratively refines the over-approximation by
filtering out sets of sets that have been shown not to be included in
$\mbb{H}$,
i.e.\ the so-called counterexamples.

\cref{alg:mono-fmp} summarizes the proposed approach\footnote{%
Arguments can either represent actual arguments or some
parameterization; these are separated by a semi-colon.}.
Also,~\cref{alg:newposcl,alg:newnegcl} provide supporting functions.
(For simplicity, the function calls
of~\cref{alg:newposcl,alg:newnegcl} show the arguments, but not the
parameterizations.)
\cref{alg:mono-fmp} iteratively uses an NP oracle (in fact a SAT
solver) to pick (or \emph{guess}) a subset $\fml{P}$ of $\fml{F}$,
such that any previously picked set is not repeated. Since we are
interested in feature $t$, we enforce that the picked set must include
$t$. (This step is shown in lines~\ref{alg:mnfmp:ln03}
to~\ref{alg:mnfmp:ln06}.)
Now, the features not in $\fml{P}$ are deemed universal, and so we
need to account for the range of possible values that these universal
features can take. For that, we update lower and upper bounds on the
predicted classes. For the features in $\fml{P}$ we must use the
values dictated by $\mbf{v}$. (This is shown in
lines~\ref{alg:mnfmp:ln07} and~\ref{alg:mnfmp:ln08}, and it is sound
to do because we have monotonicity of prediction.)
If the lower and upper bounds differ, then the picked set is not even
a weak AXp, and so we can safely remove it from further consideration.
This is achieved by enforcing that at least one of the non-picked
elements is picked in the future.
%Why? Because we want to find a set that is at least a weak AXp, and the set we picked is not one. 
(As can be observed $\fml{H}$ is updated with a positive clause that captures
this constraint, as shown in line~\ref{alg:mnfmp:ln10}.)
If the lower and upper bounds do not differ (i.e.\ we picked a weak
AXp), and if by allowing $t$ to take any value causes the bounds to
differ, then we know that any AXp in $\fml{P}$ must include $t$, and
so the algorithm reports $\fml{P}$ as a weak AXp that is
\emph{guaranteed} to be included in $\mbb{H}$. (This is shown in
line~\ref{alg:mnfmp:ln12}.)
It should be noted that $\fml{P}$ is not necessarily an AXp. However,
by~\cref{prop:fmpnp2}, $\fml{P}$ is guaranteed to be a weak AXp such
that \emph{any} of the AXp's contained in $\fml{P}$ \emph{must}
include feature $t$. From~\cite{msgcin-icml21}, we know that we can
extract an AXp from a weak AXp in polynomial time, and in this case we
are guaranteed to always pick one that includes $t$.
Finally, the last case is when allowing $t$ to take any value does not
cause the lower and upper bounds to change. This means we picked a set
$\fml{P}$ that is a weak AXp, but not all AXp's in $\fml{P}$ include
the target feature $t$ (again due to~\cref{prop:fmpnp2}). As a result,
we must prevent the same weak AXp from being re-picked. This is
achieved by requiring that at least one of the picked features not 
be picked again in the feature set. (This is shown in
line~\ref{alg:mnfmp:ln13}. As can be observed, $\fml{H}$ is updated
with a negative clause that captures this constraint.)

\begin{algorithm}[t]
  \scalebox{0.95}{
    \begin{minipage}[t]{\textwidth}
      %
\begin{comment}
\hspace*{\algorithmicindent}
\textbf{Input}: {Feature Set $\fml{F}$, Monotonic Classifier $\kappa$, Instance $\mbf{v}$, Target feature $t$}
\begin{algorithmic}[1]
%
  \Procedure{$\mathsf{Deciding FMP}$}{$\fml{F}, \kappa, \mbf{v},t$}
  \State{
    \label{alg:mnfmp:ln01}$\fml{H}\gets\{\emptyset\}$}
  \Repeat\label{alg:mnfmp:ln02}
  \State{\label{alg:mnfmp:ln03} $(\outc,\mbf{s})\gets\SAT(\fml{H}, \neg s_t)$}
  %
  \If{\label{alg:mnfmp:ln04} $\outc=\True$}
  \State{\label{alg:mnfmp:ln05}$\fml{P}\gets\{i\in\fml{F}\,|\,s_i=1\}$}
  \State{\label{alg:mnfmp:ln06}$\fml{D}\gets\{i\in\fml{F}\land i \neq t\,|\,s_i=0\}$}
  \State{\label{alg:mnfmp:ln07}$\mbf{v}_L \gets (v_{L_1}, \dots, v_{L_N}), \text{s.t.}~ v_{L_i} \gets \text{ite}(s_i, v_i, \lambda(i))$}
  \State{\label{alg:mnfmp:ln08}$\mbf{v}_U \gets (v_{U_1}, \dots, v_{U_N}), \text{s.t.}~ v_{U_i} \gets \text{ite}(s_i, v_i, \mu(i))$}
  %
  \If{\label{alg:mnfmp:ln09}$\kappa(\mbf{v}_L) = \kappa(\mbf{v}_U)$}
  \State{\label{alg:mnfmp:ln10}$\fml{H}\gets\fml{H}\cup\{(\lor_{i\in{\fml{P}}}\neg s_i)\}$}
  \Else
  \If{\label{alg:mnfmp:ln11}$\kappa(\mbf{v}_L[v_{L_t} \gets v_t]) = \kappa(\mbf{v}_U[v_{U_t} \gets v_t])$}
  \State{\label{alg:mnfmp:ln12}$\prtwaxp(\fml{P}\cup\{t\})$}
  \State{\Return}
  \EndIf
  \State{\label{alg:mnfmp:ln13}$\fml{H}\gets\fml{H}\cup\{(\lor_{i\in{\fml{D}}}s_i)\}$}
  \EndIf
  %
  \EndIf
  \Until{\label{alg:mnfmp:ln14}$\outc=\False$}
  \EndProcedure
  %
\end{algorithmic}
%
\end{comment}
%
%
\hspace*{\algorithmicindent}
\textbf{Input}: {Instance $\mbf{v}$}, Target feature $t$; Feature Set $\fml{F}$, Monotonic Classifier $\kappa$
\begin{algorithmic}[1]
  %
  %\Procedure{$\mathsf{Decide FMP}$}{$\mbf{v}, t; \fml{F}, \kappa$}
  \Function{$\mathsf{Decide Relevant}$}{$\mbf{v}, t; \fml{F}, \kappa$}
  %\LineComment{$\fml{H}$ abstracts the sets that are not weak AXp's or do not contain AXp's that include $t$}
  %\LineComment{$\fml{H}$ overapproximates the subsets of $\fml{F}$
  %  that do not contain an AXp containing $t$}
  \State{\label{alg:mnfmp:ln01}$\fml{H}\gets\emptyset$}
  \Comment{$\fml{H}$ overapproximates $\mbb{H}$}
  \Repeat\label{alg:mnfmp:ln02}
  \State{\label{alg:mnfmp:ln03}$(\outc,\mbf{s})\gets\SAT(\fml{H},s_t)$}
  \Comment{Pick candidate weak AXp containing $t$}
  \If{\label{alg:mnfmp:ln04} $\outc=\True$}
  \State{\label{alg:mnfmp:ln05}$\fml{P}\gets\{i\in\fml{F}\,|\,s_i=1\}$}
  \Comment{$\fml{P}$ is the candidate weak AXp, and $t\in\fml{P}$}
  \State{\label{alg:mnfmp:ln06}$\fml{D}\gets\{i\in\fml{F}\,|\,s_i=0\}$}
  \Comment{$\fml{D}$ contains the features not included in $\fml{P}$}
  \State{\label{alg:mnfmp:ln07}$\mbf{v}_L \gets (v_{L_1}, \dots, v_{L_N}), \text{s.t.}~ v_{L_i} \gets \text{ITE}(s_i, v_i, \lambda(i))$}
  \Comment{$\mbf{v}_L$: LB} %lower bound given $\fml{P}$;on possible predictions
  \State{\label{alg:mnfmp:ln08}$\mbf{v}_U \gets (v_{U_1}, \dots, v_{U_N}), \text{s.t.}~ v_{U_i} \gets \text{ITE}(s_i, v_i, \mu(i))$}
  \Comment{$\mbf{v}_U$: UB} %upper bound given $\fml{P}$;on possible predictions
  \If{\label{alg:mnfmp:ln09}$\kappa(\mbf{v}_L) \neq \kappa(\mbf{v}_U)$}
  \Comment{More than one value possible?}
  %\State{\label{alg:mnfmp:ln10}$\fml{H}\gets\fml{H}\cup\{(\lor_{i\in{\fml{D}}}s_i)\}$}
  \State{\label{alg:mnfmp:ln10}$\fml{H}\gets\fml{H}\cup\newpcl(\fml{D},t)$}
  \Comment{$\fml{P}$ is \emph{not} a weak AXp; block set}
  \Else
  \Comment{$\fml{P}$ is a weak AXp}
  \If{\label{alg:mnfmp:ln11}$\kappa(\mbf{v}_L[v_{L_t} \gets \lambda(t)]) \neq \kappa(\mbf{v}_U[v_{U_t} \gets \mu(t)])$}
  \Comment{$t$ needed?}
  \State{\label{alg:mnfmp:ln12}$\prtwaxp(\fml{P})$}
  \Comment{$t$ is included in any AXp $\fml{X}\subseteq\fml{P}$}
  \State{\Return \True}
  \EndIf
  %\State{\label{alg:mnfmp:ln13}$\fml{H}\gets\fml{H}\cup\{(\lor_{i\in{\fml{P}} \land i \neq t}\neg s_i)\}$}
  \State{\label{alg:mnfmp:ln13}$\fml{H}\gets\fml{H}\cup\newncl(\fml{P},t)$}
  \Comment{$t$ unneeded; block set} %not required
  \EndIf
  \EndIf
  \Until{\label{alg:mnfmp:ln14}$\outc=\False$}
  \State{\Return \False}
  \Comment{If $\fml{H}$ becomes inconsistent, then \emph{no} AXp contains $t$}
  %, i.e.\ the overapproximatation is the empty set
  %\EndProcedure
  \EndFunction
\end{algorithmic}

    \end{minipage}
  }
  \caption{Deciding feature relevancy for a monotonic classifier} \label{alg:mono-fmp}
\end{algorithm}

As can be concluded from~\cref{alg:mono-fmp} and from the discussion
above, \cref{prop:fmpnp2} is essential to enable us to use at most two
classification queries per iteration of the algorithm. If we were to
use~\cref{prop:fmpnp} instead, then the number of classification
queries would be significantly larger.

\begin{figure*}[t]%[ht]
  \begin{minipage}{0.485\textwidth} %[t]
    \begin{algorithm}[H]
      \scalebox{0.8725}{
        \begin{minipage}[t]{\textwidth}
          \input{./algs/newposcl}
        \end{minipage}
      }
      \caption{Create new pos.~clause} \label{alg:newposcl}
    \end{algorithm}
  \end{minipage}
  \begin{minipage}{0.0\textwidth} %[t]
    ~
  \end{minipage}
  \begin{minipage}{0.5075\textwidth} %[t]
    \begin{algorithm}[H]
      \scalebox{0.8725}{
        \begin{minipage}[t]{\textwidth}
          \input{./algs/newnegcl}
        \end{minipage}
      }
      \caption{Create new neg.~clause} \label{alg:newnegcl}
    \end{algorithm}
  \end{minipage}
\end{figure*}

\begin{example}
  We consider the monotonic classifier of~\cref{fig:runex02}, with
  instance $(\mbf{v},c)=((1,1,1,1),1)$.
  \cref{tab:rex02a} summarizes a possible execution of the algorithm
  when $t=4$.
  Similarly, \cref{tab:rex02b} summarizes a possible execution of the
  algorithm when $t=1$.
  (As with the current implementation, and for both examples, the
  creation of clauses uses no optimizations.)
  In general, different executions will be  determined by the models
  returned by the SAT solver. 
\end{example}

\begin{table}[t]
  \caption{Example algorithm execution for $t=4$} \label{tab:rex02a}
  \hspace*{-0.225cm}
    \scalebox{0.9175}{
      \renewcommand{\tabcolsep}{0.275em}
      \begin{tabular}{C{1.45cm}C{1.275cm}C{1.275cm}C{0.95cm}C{0.95cm}C{2.75cm}C{2.25cm}C{0.675cm}}
        \toprule
        $\mbf{s}$ & $\fml{P}$ & $\fml{D}$ & $\kappa(\mbf{v}_L)$ & $\kappa(\mbf{v}_U)$ &
        Decision & New clause & Line
        \\ \toprule
        $(0,0,0,1)$ & $\{4\}$ & $\{1,2,3\}$ & 0 & 1 &
        New pos clause & $(s_1\lor{s_2}\lor{s_3})$ & \ref{alg:mnfmp:ln10}
        \\ \midrule
        $(1,0,0,1)$ & $\{1,4\}$ & $\{2,3\}$ & 0 & 1 &
        New pos clause & $({s_2}\lor{s_3})$ & \ref{alg:mnfmp:ln10}
        \\ \midrule
        $(1,1,0,1)$ & $\{1,2,4\}$ & $\{3\}$ & 1 & 1 &
        New neg clause & $(\neg{s_1}\lor\neg{s_2})$ & \ref{alg:mnfmp:ln13} 
        \\ \midrule
        $(1,0,1,1)$ & $\{1,3,4\}$ & $\{2\}$ & 1 & 1 &
        New neg clause & $(\neg{s_1}\lor\neg{s_3})$ & \ref{alg:mnfmp:ln13} 
        \\ \midrule
        $(0,1,1,1)$ & $\{2,3,4\}$ & $\{1\}$ & 1 & 1 &
        New pos clause & $({s_1})$ & \ref{alg:mnfmp:ln10} 
        \\ \midrule
        --- & --- & --- & -- & -- &
        $\fml{H}$ inconsistent & -- & \ref{alg:mnfmp:ln14}
        \\ \bottomrule
      \end{tabular}
    }
\end{table}

\begin{table}[t]
  \caption{Example algorithm execution for $t=1$} \label{tab:rex02b}
  \hspace*{-0.225cm}
    \scalebox{0.9175}{
      \renewcommand{\tabcolsep}{0.275em}
      \begin{tabular}{C{1.45cm}C{1.275cm}C{1.275cm}C{0.95cm}C{0.95cm}C{2.75cm}C{2.25cm}C{0.675cm}}
        \toprule
        $\mbf{s}$ & $\fml{P}$ & $\fml{D}$ & $\kappa(\mbf{v}_L)$ &
        $\kappa(\mbf{v}_U)$ &
        Decision & New clause & Line
        \\ \toprule
        $(1,0,0,0)$ & $\{1\}$ & $\{2,3,4\}$ & 0 & 1 &
        New pos clause & $(s_2\lor{s_3}\lor{s_4})$ & \ref{alg:mnfmp:ln10}
        \\ \midrule
        $(1,1,0,0)$ & $\{1,2\}$ & $\{3,4\}$ & 1 & 1 &
        Weak AXp: $\{1,2\}$ & -- & \ref{alg:mnfmp:ln12}
        \\ \bottomrule
      \end{tabular}
    }
\end{table}

With respect to the clauses that are added to $\fml{H}$ at each step,
as shown in~\cref{alg:newposcl,alg:newnegcl}, one can envision
optimizations (shown lines~\ref{alg:newncl:ln02}
to~\ref{alg:newncl:ln06} in both algorithms)
that heuristically aim at removing features from the given sets, and
so produce shorter (and so logically stronger) clauses.
The insight is that any feature, which can be deemed irrelevant for
the condition used for constructing the clause, can be safely
removed from the set.
(In practice, our experiments show that the time running the
classifier is far larger than the time spent using the NP oracle to
guess sets. Thus,
we opted to use the simplest approach for constructing the clauses,
and so reduce the number of classification queries.)

Given the above discussion, we can conclude that the proposed
algorithm is sound, complete and terminating for deciding
feature relevancy for monotonic classifiers.
(The proof is straightforward, and it is omitted for the sake of
brevity.)

\begin{prop} \label{prop:algok}
  For a monotonic classifier $\mbb{C}$, defined on set of features
  $\fml{F}$, with $\kappa$ mapping $\mbb{F}$ to $\fml{K}$, and an
  instance $(\mbf{v},c)$, $\mbf{v}\in\mbb{F}$, $c\in\fml{K}$, and a
  target feature $t\in\fml{F}$,~\cref{alg:mono-fmp} returns a set
  $\fml{P}\subseteq\fml{F}$ iff $\fml{P}$ is a weak AXp for
  $(\mbf{v},c)$, with the property that any AXp
  $\fml{X}\subseteq\fml{P}$ is such that $t\in\fml{X}$ (i.e. $\fml{P}$ is a witness for the relevancy of $t$).
\end{prop}

\section{Experimental Results} \label{sec:res}
This section reports the experimental results on FRP
for the d-DNNF and monotonic classifiers.
The goal is to show that FRP is practically feasible.
We opt not to include experiments for FNP as the complexity of FNP is in P.
Besides, to the best of our knowledges, there is no baseline to compare with.
The experiments were performed on a MacBook Pro with a 6-Core Intel
Core~i7 2.6~GHz processor with 16~GByte RAM, running macOS Monterey.

\paragraph{d-DNNF classifiers.}
For d-DNNFs, we pick its subset SDDs as our target classifier.
SDDs support polynomial time negation,
so given a SDD $\mbb{C}$, one can obtain its negation $\neg \mbb{C}$ efficiently.

\paragraph{Monotonic classifiers.}
For monotonic classifiers, we consider the Deep Lattice Network (DLN)~\cite{gupta-nips17} as our target classifier
\footnote{\url{https://github.com/tensorflow/lattice}}.
Since our approach for monotonic classifier is model-agnostic, it could also be used with other
approaches for learning monotonic
classifiers~\cite{louppe-nips19,liu-nips20} including Min-Max
Network~\cite{sill-nips97,velikova-tnn10} and
COMET~\cite{vandenbroeck-nips20}. 

\paragraph{Prototype implementation.}
Prototype implementations of the proposed
approaches were implemented in Python~\footnote{\url{https://github.com/XuanxiangHuang/frp-experiment}}.
The PySAT toolkit~\footnote{\url{https://github.com/pysathq/pysat}} was used for propositional encodings.
Besides, PySAT invokes the Glucose 4~\footnote{\url{https://www.labri.fr/perso/lsimon/glucose/}} SAT solver to pick a weak AXp candidate.
SDDs were loaded by using the 
PySDD~\footnote{\url{https://github.com/wannesm/PySDD}}package.

\paragraph{Benchmarks \& training.}
For SDDs,
we selected 11 datasets from
Density Estimation Benchmark Datasets\footnote{\url{https://github.com/UCLA-StarAI/Density-Estimation-Datasets}}
~\cite{lowd2010learning,van2012markov,larochelle2011neural}.
11 datasets were used to learn SDD using LearnSDD\footnote{\url{https://github.com/ML-KULeuven/LearnSDD}}
~\cite{bekker2015tractable}
(with parameter \textit{maxEdges=20000}).
The obtained SDDs were used as binary classifiers.
%%%
For DLNs,
we selected 5 publicly available datasets:
\emph{australian} (aus), \emph{breast\_cancer} (b.c.), \emph{heart\_c}, \emph{nursery}\footnote{\url{https://epistasislab.github.io/pmlb/index.html}}
~\cite{Olson2017PMLB}
and \emph{pima}\footnote{\url{https://sci2s.ugr.es/keel/dataset.php?cod=21}}
~\cite{alcala2011keel}.
We used the three-layer DLN architecture: 
Calibrators $\to$ Random Ensemble of Lattices $\to$ Linear Layer.
All calibrators for all models used a fixed number of 20 keypoints.
And the size of all lattices was set to 3.
\begin{table}[t]
\begin{center}
\caption{Solving FRP for SDDs.
Sub-Columns Avg. \#var and Avg. \#cls show, respectively,
the average number of variables and clauses in a CNF encoding.
Column Runtime reports maximum and average time in seconds
for deciding FRP.
}
\label{tab:sdd}
\setlength{\tabcolsep}{2pt}

\begin{tabular}{cccccccc}
\toprule[1.2pt]
\multirow{2}{*}{Dataset} & \multicolumn{2}{c}{SDD} & \multirow{2}{*}{\%Y} & \multicolumn{2}{c}{CNF} & \multicolumn{2}{c}{Runtime (s)} \\
\cmidrule[0.5pt]{2-3}
\cmidrule[0.5pt]{5-8}
                      & \#Features        & \#Nodes        &                        & Avg. \#var & Avg. \#cls & Max             & Avg.          \\
\midrule[0.8pt]
Accidents             & 415        & 8863       & 97                     & 26513      & 78276      & 56.4            & 3.5           \\
Audio                 & 272        & 7224       & 88                     & 31148      & 100972     & 663.1           & 22.0          \\
DNA                   & 513        & 8570       & 91                     & 29155      & 91288      & 86.3            & 11.0          \\
Jester                & 254        & 7857       & 85                     & 35998      & 121508     & 362.1           & 22.7          \\
KDD                   & 306        & 8109       & 99                     & 26402      & 83480      & 111.2           & 2.8           \\
Mushrooms             & 248        & 7096       & 91                     & 23874      & 82112      & 266.3           & 15.8          \\
Netflix               & 292        & 7039       & 94                     & 25520      & 83324      & 105.7           & 4.2           \\
NLTCS                 & 183        & 6661       & 100                    & 19817      & 58494      & 1.4             & 0.5           \\
Plants                & 244        & 6724       & 97                     & 25356      & 84782      & 950.7           & 20.6          \\
RCV-1                 & 410        & 9472       & 90                     & 33438      & 102500     & 153.6           & 11.2          \\
Retail                & 341        & 3704       & 87                     & 10601      & 28342      & 1.8             & 1.1           \\
\bottomrule[1.2pt]
\end{tabular}
\end{center}
\end{table}

\paragraph{Results for SDDs.}
For each SDD, 100 test instances were randomly generated.
All tested instances have prediction $0$.
(We didn't pick instances predicted to class $1$
as this requires the compilation of a new classifier which may have different size).
Besides, for each instance, we randomly picked a feature
appearing in the model.
Hence for each SDD, we solved 100 queries.
Table \ref{tab:sdd} summarizes the results.
It can be observed that the number of nodes of the tested SDD
is in the range of 3704 and 9472, and the number of features of tested SDD
is in the range of 183 and 513.
Besides, the percentage of examples for which the answer is
Y (i.e.\ target feature is in some AXp) ranges from 85\% to 100\%.
Regarding the runtime, the largest running time for solving one query can exceed 15 minutes.
But the average running time to solve a query is less than 25 seconds,
this highlights the scalability of the proposed encoding.

\paragraph{Results for DLNs.}
\begin{table}[t]
\begin{center}
\caption{Solving FRP for DLN.
%
%Column Y\% shows the percentage of answering `Yes' to the queries.
%
Column Runtime reports maximum and average time in seconds
for deciding FRP.
Column SAT Time (resp. $\kappa(\mathbf{v})$ Time) reports maximum and average time in seconds
for SAT solver (resp. calling DLN's predict function) to decide FRP.
Column SAT Calls (resp. $\kappa(\mathbf{v})$ Calls) reports maximum and average number of calls
to the SAT solver (resp. to the DLN's predict function) to decide FRP.
}
\label{tab:mono-dln}

\setlength{\tabcolsep}{2pt}
\begin{tabular}{lrrrrrrrrrrrrrr}
\toprule[1.2pt]
\multicolumn{1}{c}{\multirow{2}{*}{Dataset}} & \multicolumn{1}{c}{\multirow{2}{*}{\%Y}} & \multicolumn{2}{c|}{Runtime (s)} & \multicolumn{2}{c|}{SAT Time} & \multicolumn{2}{c|}{SAT Calls} & \multicolumn{2}{c|}{$\kappa(\mathbf{v})$ Time} & \multicolumn{2}{c}{$\kappa(\mathbf{v})$ Calls} & \multirow{2}{*}{$\frac{\kappa(\mathbf{v}) \text{Time}}{\text{Runtime}}$} \\
\cmidrule[0.5pt]{3-12}
\multicolumn{1}{r}{}                      & \multicolumn{1}{r}{}                                 & Max             & Avg.           & Max           & Avg.         & Max             & Avg.           & Max            & Avg.         & Max             & Avg.           \\
\midrule[0.8pt]
aus & 61 & 40.4 & 8.31 & 0.02 & 0.01 & 291 & 65 & 40.0 & 8.15 & 424 & 98 & 97.8\% \\
b.c. & 45 & 5.4 & 1.93 & 0.00 & 0.00 & 53 & 20 & 5.3 & 1.89 & 78 & 30 & 98.0\% \\
heart\_c & 35 & 31.5 & 6.67 & 0.02 & 0.00 & 171 & 54 & 31.1 & 6.52 & 249 & 80 & 97.7\% \\
nursery  & 45 & 4.3 & 1.77 & 0.00 & 0.00 & 31 & 13 & 4.3 & 1.75 & 73 & 30 & 98.6\% \\
pima & 74 & 3.7 & 1.41 & 0.00 & 0.00 & 33 & 13 & 3.7 & 1.39 & 47 & 22 & 98.4\% \\
\bottomrule[1.2pt]
\end{tabular}
%%%
\end{center}
\end{table}

For each DLN, we randomly picked 200 tested instances,
and for each tested instance, we randomly pick a feature.
Hence for each DLN, we solved 200 queries.
\cref{tab:mono-dln} summarizes the results.
The use of a SAT solver has a negligible
contribution to the running time. Indeed, for all the examples shown,
at least 97\% of the running time is spent running the
classifier. This should be unsurprising, since the number of the
iterations of~\cref{alg:mono-fmp} never exceeds a few hundred.
(The fraction of a second reported in some cases should be divided by the
number of calls to the SAT solver; hence the time spent in each call
to the SAT solver is indeed negligible.)
As can be observed, the percentage of examples for which the answer is
Y (i.e.\ target feature is in some AXp and the algorithm returns
\textbf{true}) ranges from 35\% to 74\%.
There is no apparent correlation between the percentage of Y answers and the number of
iterations.
The large number of queries accounts for the number of times the DLN
is queried by~\cref{alg:mono-fmp}, but it also accounts for the number
of times the DLN is queried for extracting an AXp from set $\fml{P}$
(i.e.\ the witness) when the algorithm's answer is \textbf{true}.
A loose upper bound on the number of queries to the classifier is
$4\times{\tn{NS}}+2\times|\fml{F}|$, where $\tn{NS}$ is the number of
SAT calls, and $|\fml{F}|$ is the number of features. Each iteration
of \cref{alg:mono-fmp} can require at most 4 queries to the
classifier. After reporting $\fml{P}$, at most 2 queries per feature 
will be required to extract the AXp (see~\cref{ssec:fxai}). As can be
observed this loose upper bound is respected by the reported results.
%%%

% ...

\section{Related Work} \label{sec:relw}

The problems of necessity and relevancy have been studied in
logic-based abduction since the early
90s~\cite{gottlob-ese90,selman-aaai90,gottlob-jacm95}. However, this
earlier work did not consider the classes of (classifier) functions
that are considered in this paper.

There has been recent work on explainability
queries~\cite{marquis-kr20,marquis-kr21,hiims-kr21}.
Some of these queries can be related with feature relevancy and
necessity. For example, relevancy and necessity have been studied with
respect to a target class~\cite{marquis-kr20,marquis-kr21}, in
contrast with our approach that studies a concrete instance, and so
can be naturally related with earlier work on abduction.
Recent work\cite{hiims-kr21} studied feature relevancy under the name feature membership,
but neither d-DNNF nor monotonic classifiers were discussed.
Moreover, \cite{hiims-kr21} only proved the hardness of deciding feature relevancy for DNF and DT classifiers
and did not discuss the feature necessity problem.
The results presented in this paper complement this work.
Besides, the complexity results of FRP and FNP in this paper
also complement the recent work~\cite{msi-aaai22} which summarizes the progress of formal explanations.
\cite{inms-aaai19} focused on the computation of one arbitrary AXp and one smallest AXp,
which is orthogonal to our work. Computing one AXp does not guarantee that either FRP or FNP is decided,
since the target feature $t$ may not appear in the computed AXp.
\cite{msgcin-icml21} studied the computation of one formal explanation and the enumeration of formal explanations
in the case study of monotonic classifiers.
However, neither FRP or FNP were identified and studied.

\section{Conclusions} \label{sec:conc}

This paper studies the problems of feature necessity and relevancy in
the context of formal explanations of ML classifiers.
The paper proves several complexity results, some related with
necessity, but most related with relevancy. Furthermore, the paper
proposes two different approaches for solving relevancy for two
families of classifiers, namely classifiers represented with the
d-DNNF propositional language, and monotonic classifiers.
The experimental results confirm the practical scalability of the
proposed algorithms.
Future work will seek to prove hardness results for the families of
classifiers for which hardness is yet unknown.

\subsubsection{Acknowledgements}
This work was supported by the AI Interdisciplinary Institute ANITI,
funded by the French program ``Investing for the Future -- PIA3''
under Grant agreement no.\ ANR-19-PI3A-0004, and by the H2020-ICT38
project COALA ``Cognitive Assisted agile manufacturing for a Labor
force supported by trustworthy Artificial intelligence'',
and funded by the Spanish Ministry of Science and Innovation 
(MICIN) under project PID2019-111544GB-C22, and by a María Zambrano
fellowship and a Requalification fellowship financed by Ministerio 
de Universidades of Spain and by European Union -- NextGenerationEU.
  % Uncomment for non-anonymous version.
%
%\clearpage
%\bibliographystyle{splncs04}
%\bibliography{refs,team}

% RequiredL: \usepackage{etoolbox}
%\providetoggle{mkbbl}
\newtoggle{mkbbl}
% Contents if using bibtex: "\settoggle{mkbbl}{true}"
% Contents if inputing pre-generated file: "\settoggle{mkbbl}{false}"

\settoggle{mkbbl}{false}
 % file is automatically generated

% ---- Bibliography ----
%%\cleardoublepage %% TENTATIVE, and required if bibliography starts page...
\addcontentsline{toc}{section}{References}
\vskip 0.2in
% For arxix paper production, and since arXiv does not allow for
% bibtex, we need to create a .bbl file to include upon submission
% to arXiv.
\iftoggle{mkbbl}{
  % Run bibtex, i.e. generate .bbl gile
  \bibliographystyle{splncs04}
  \bibliography{refs,team}

\begin{thebibliography}{10}
\providecommand{\url}[1]{\texttt{#1}}
\providecommand{\urlprefix}{URL }
\providecommand{\doi}[1]{https://doi.org/#1}

\bibitem{akers1978binary}
Akers, S.B.: Binary decision diagrams. IEEE Transactions on computers
  \textbf{27}(06),  509--516 (1978)

\bibitem{alcala2011keel}
Alcal{\'a}-Fdez, J., Fern{\'a}ndez, A., Luengo, J., Derrac, J., Garc{\'\i}a,
  S., S{\'a}nchez, L., Herrera, F.: Keel data-mining software tool: data set
  repository, integration of algorithms and experimental analysis framework.
  Journal of Multiple-Valued Logic \& Soft Computing  \textbf{17} (2011),
  \url{https://sci2s.ugr.es/keel/dataset.php?cod=21}

\bibitem{amgoud-ijcai22}
Amgoud, L., Ben{-}Naim, J.: Axiomatic foundations of explainability. In: IJCAI.
  pp. 636--642 (2022)

\bibitem{barcelo-nips21}
Arenas, M., Baez, D., Barcel{\'{o}}, P., P{\'{e}}rez, J., Subercaseaux, B.:
  Foundations of symbolic languages for model interpretability. In: NeurIPS
  (2021)

\bibitem{barcelo-corr22}
Arenas, M., Barcel{\'{o}}, P., Romero, M., Subercaseaux, B.: On computing
  probabilistic explanations for decision trees. CoRR  \textbf{abs/2207.12213}
  (2022). \doi{10.48550/arXiv.2207.12213},
  \url{https://doi.org/10.48550/arXiv.2207.12213}

\bibitem{arora-bk09}
Arora, S., Barak, B.: Computational Complexity - {A} Modern Approach. Cambridge
  University Press (2009),
  \url{http://www.cambridge.org/catalogue/catalogue.asp?isbn=9780521424264}

\bibitem{marquis-kr21}
Audemard, G., Bellart, S., Bounia, L., Koriche, F., Lagniez, J., Marquis, P.:
  On the computational intelligibility of boolean classifiers. In: KR. pp.
  74--86 (2021)

\bibitem{marquis-kr20}
Audemard, G., Koriche, F., Marquis, P.: On tractable {XAI} queries based on
  compiled representations. In: KR. pp. 838--849 (2020)

\bibitem{muller-plosone15}
Bach, S., Binder, A., Montavon, G., Klauschen, F., M{\"u}ller, K.R., Samek, W.:
  On pixel-wise explanations for non-linear classifier decisions by layer-wise
  relevance propagation. PloS one  \textbf{10}(7),  e0130140 (2015)

\bibitem{barcelo-nips20}
Barcel{\'{o}}, P., Monet, M., P{\'{e}}rez, J., Subercaseaux, B.: Model
  interpretability through the lens of computational complexity. In: NeurIPS
  (2020)

\bibitem{bekker2015tractable}
Bekker, J., Davis, J., Choi, A., Darwiche, A., den Broeck, G.V.: Tractable
  learning for complex probability queries. In: NeurIPS. pp. 2242--2250 (2015),
  \url{https://github.com/ML-KULeuven/LearnSDD}

\bibitem{bengio-cacm21}
Bengio, Y., LeCun, Y., Hinton, G.E.: Deep learning for {AI}. Commun. {ACM}
  \textbf{64}(7),  58--65 (2021), \url{https://doi.org/10.1145/3448250}

\bibitem{sat-hbk21}
Biere, A., Heule, M., van Maaren, H., Walsh, T. (eds.): Handbook of
  Satisfiability - Second Edition, Frontiers in Artificial Intelligence and
  Applications, vol.~336. {IOS} Press (2021),
  \url{https://doi.org/10.3233/FAIA336}

\bibitem{tan-nips21}
Blanc, G., Lange, J., Tan, L.: Provably efficient, succinct, and precise
  explanations. In: NeurIPS (2021)

\bibitem{mazure-cikm21}
Boumazouza, R., Alili, F.C., Mazure, B., Tabia, K.: {ASTERYX:} {A}
  model-{A}gnostic {S}a{T}-bas{E}d app{R}oach for s{Y}mbolic and score-based
  e{X}planations. In: CIKM. pp. 120--129 (2021)

\bibitem{brayton1984logic}
Brayton, R.K., Hachtel, G.D., McMullen, C., Sangiovanni-Vincentelli, A.: Logic
  minimization algorithms for VLSI synthesis, vol.~2. Springer Science \&
  Business Media (1984)

\bibitem{breiman2001random}
Breiman, L.: Random forests. Mach. Learn.  \textbf{45}(1),  5--32 (2001).
  \doi{10.1023/A:1010933404324}, \url{https://doi.org/10.1023/A:1010933404324}

\bibitem{breiman-bk84}
Breiman, L., Friedman, J.H., Olshen, R.A., Stone, C.J.: Classification and
  Regression Trees. Wadsworth (1984)

\bibitem{clark1991rule}
Clark, P., Boswell, R.: Rule induction with cn2: Some recent improvements. In:
  European Working Session on Learning. pp. 151--163. Springer (1991)

\bibitem{clarke-jacm03}
Clarke, E.M., Grumberg, O., Jha, S., Lu, Y., Veith, H.: Counterexample-guided
  abstraction refinement for symbolic model checking. J. {ACM}  \textbf{50}(5),
   752--794 (2003). \doi{10.1145/876638.876643},
  \url{https://doi.org/10.1145/876638.876643}

\bibitem{velikova-tnn10}
Daniels, H., Velikova, M.: Monotone and partially monotone neural networks.
  {IEEE} Trans. Neural Networks  \textbf{21}(6),  906--917 (2010)

\bibitem{darwiche-ijcai11}
Darwiche, A.: {SDD:} {A} new canonical representation of propositional
  knowledge bases. In: IJCAI. pp. 819--826 (2011)

\bibitem{darwiche-jair02}
Darwiche, A., Marquis, P.: A knowledge compilation map. J. Artif. Intell. Res.
  \textbf{17},  229--264 (2002). \doi{10.1613/jair.989}

\bibitem{darwiche-jair21}
Darwiche, A., Marquis, P.: On quantifying literals in boolean logic and its
  applications to explainable {AI}. J. Artif. Intell. Res.  \textbf{72},
  285--328 (2021)

\bibitem{gottlob-jacm95}
Eiter, T., Gottlob, G.: The complexity of logic-based abduction. J. {ACM}
  \textbf{42}(1),  3--42 (1995), \url{https://doi.org/10.1145/200836.200838}

\bibitem{gupta-nips16}
Fard, M.M., Canini, K.R., Cotter, A., Pfeifer, J., Gupta, M.R.: Fast and
  flexible monotonic functions with ensembles of lattices. In: NeurIPS. pp.
  2919--2927 (2016)

\bibitem{leite-kr22}
Ferreira, J., de~Sousa~Ribeiro, M., Gon{\c{c}}alves, R., Leite, J.: Looking
  inside the black-box: Logic-based explanations for neural networks. In: KR.
  p. 432–442 (2022)

\bibitem{flach-bk12}
Flach, P.A.: Machine Learning - The Art and Science of Algorithms that Make
  Sense of Data. CUP (2012)

\bibitem{friedman2001greedy}
Friedman, J.H.: Greedy function approximation: a gradient boosting machine.
  Annals of statistics pp. 1189--1232 (2001)

\bibitem{gottlob-ese90}
Friedrich, G., Gottlob, G., Nejdl, W.: Hypothesis classification, abductive
  diagnosis and therapy. In: ESE. pp. 69--78 (1990)

\bibitem{gergov1992efficient}
Gergov, J., Meinel, C.: Efficient boolean manipulation with {OBDD}'s can be
  extended to {FBDD}'s. IEEE Transactions on Computers  \textbf{43}(10),
  1197--1209 (1994). \doi{10.1109/12.324545}

\bibitem{bengio-bk16}
Goodfellow, I.J., Bengio, Y., Courville, A.C.: Deep Learning. Adaptive
  computation and machine learning, {MIT} Press (2016),
  \url{http://www.deeplearningbook.org/}

\bibitem{rubin-aaai22}
Gorji, N., Rubin, S.: Sufficient reasons for classifier decisions in the
  presence of domain constraints. In: {AAAI} (February 2022)

\bibitem{van2012markov}
Haaren, J.V., Davis, J.: Markov network structure learning: {A} randomized
  feature generation approach. In: AAAI (2012)

\bibitem{hiicams-aaai22}
Huang, X., Izza, Y., Ignatiev, A., Cooper, M.C., Asher, N., Marques{-}Silva,
  J.: Tractable explanations for {d-DNNF} classifiers. In: AAAI. pp. 5719--5728
  (2022)

\bibitem{hiims-kr21}
Huang, X., Izza, Y., Ignatiev, A., Marques{-}Silva, J.: On efficiently
  explaining graph-based classifiers. In: KR. pp. 356--367 (2021)

\bibitem{iisms-aaai22}
Ignatiev, A., Izza, Y., Stuckey, P.J., Marques{-}Silva, J.: Using {MaxSAT} for
  efficient explanations of tree ensembles. In: AAAI. pp. 3776--3785 (2022)

\bibitem{ims-sat21}
Ignatiev, A., Marques{-}Silva, J.: {SAT}-based rigorous explanations for
  decision lists. In: SAT. pp. 251--269 (2021)

\bibitem{inams-aiia20}
Ignatiev, A., Narodytska, N., Asher, N., Marques{-}Silva, J.: From contrastive
  to abductive explanations and back again. In: AIxIA. pp. 335--355 (2020)

\bibitem{inms-aaai19}
Ignatiev, A., Narodytska, N., Marques{-}Silva, J.: Abduction-based explanations
  for machine learning models. In: AAAI. pp. 1511--1519 (2019)

\bibitem{ipnms-ijcar18}
Ignatiev, A., Pereira, F., Narodytska, N., Marques{-}Silva, J.: A {SAT}-based
  approach to learn explainable decision sets. In: IJCAR. pp. 627--645 (2018)

\bibitem{iims-jair22}
Izza, Y., Ignatiev, A., Marques{-}Silva, J.: On tackling explanation redundancy
  in decision trees. J. Artif. Intell. Res.  \textbf{75},  261--321 (2022),
  \url{https://doi.org/10.1613/jair.1.13575}

\bibitem{ims-ijcai21}
Izza, Y., Marques{-}Silva, J.: On explaining random forests with {SAT}. In:
  IJCAI. pp. 2584--2591 (2021)

\bibitem{kohavi1994bottom}
Kohavi, R.: Bottom-up induction of oblivious read-once decision graphs:
  strengths and limitations. In: AAAI. pp. 613--618 (1994)

\bibitem{kohavi1996scaling}
Kohavi, R., et~al.: Scaling up the accuracy of naive-bayes classifiers: A
  decision-tree hybrid. In: Kdd. vol.~96, pp. 202--207 (1996)

\bibitem{larochelle2011neural}
Larochelle, H., Murray, I.: The neural autoregressive distribution estimator.
  In: AISTATS. pp. 29--37 (2011)

\bibitem{bengio-nature15}
LeCun, Y., Bengio, Y., Hinton, G.: Deep learning. nature  \textbf{521}(7553),
  436--444 (2015)

\bibitem{liu-nips20}
Liu, X., Han, X., Zhang, N., Liu, Q.: Certified monotonic neural networks. In:
  NeurIPS (2020)

\bibitem{lowd2010learning}
Lowd, D., Davis, J.: Learning {M}arkov network structure with decision trees.
  In: ICDM. pp. 334--343 (2010)

\bibitem{lundberg-nips17}
Lundberg, S.M., Lee, S.: A unified approach to interpreting model predictions.
  In: NeurIPS. pp. 4765--4774 (2017)

\bibitem{kwiatkowska-ijcai21}
Malfa, E.L., Michelmore, R., Zbrzezny, A.M., Paoletti, N., Kwiatkowska, M.: On
  guaranteed optimal robust explanations for {NLP} models. In: IJCAI. pp.
  2658--2665 (2021)

\bibitem{msgcin-nips20}
Marques{-}Silva, J., Gerspacher, T., Cooper, M.C., Ignatiev, A., Narodytska,
  N.: Explaining naive bayes and other linear classifiers with polynomial time
  and delay. In: NeurIPS (2020)

\bibitem{msgcin-icml21}
Marques{-}Silva, J., Gerspacher, T., Cooper, M.C., Ignatiev, A., Narodytska,
  N.: Explanations for monotonic classifiers. In: ICML. pp. 7469--7479 (2021)

\bibitem{msi-aaai22}
Marques{-}Silva, J., Ignatiev, A.: Delivering trustworthy {AI} through formal
  {XAI}. In: AAAI. pp. 12342--12350 (2022)

\bibitem{miller-aij19}
Miller, T.: Explanation in artificial intelligence: Insights from the social
  sciences. Artif. Intell.  \textbf{267},  1--38 (2019)

\bibitem{muller1995neural}
M{\"u}ller, B., Reinhardt, J., Strickland, M.T.: Neural networks: an
  introduction. Springer Science \& Business Media (1995)

\bibitem{Olson2017PMLB}
Olson, R.S., La~Cava, W., Orzechowski, P., Urbanowicz, R.J., Moore, J.H.:
  {PMLB:} a large benchmark suite for machine learning evaluation and
  comparison. BioData Mining  \textbf{10}(1), ~36 (2017),
  \url{https://epistasislab.github.io/pmlb/index.html}

\bibitem{guestrin-kdd16}
Ribeiro, M.T., Singh, S., Guestrin, C.: "{W}hy should {I} trust you?":
  Explaining the predictions of any classifier. In: KDD. pp. 1135--1144 (2016)

\bibitem{guestrin-aaai18}
Ribeiro, M.T., Singh, S., Guestrin, C.: Anchors: High-precision model-agnostic
  explanations. In: AAAI. pp. 1527--1535 (2018)

\bibitem{rivest-ml87}
Rivest, R.L.: Learning decision lists. Mach. Learn.  \textbf{2}(3),  229--246
  (1987)

\bibitem{selman-aaai90}
Selman, B., Levesque, H.J.: Abductive and default reasoning: {A} computational
  core. In: AAAI. pp. 343--348 (1990)

\bibitem{shalev-shwartz-bk14}
Shalev{-}Shwartz, S., Ben{-}David, S.: Understanding Machine Learning - From
  Theory to Algorithms. Cambridge University Press (2014)

\bibitem{darwiche-ijcai18}
Shih, A., Choi, A., Darwiche, A.: A symbolic approach to explaining bayesian
  network classifiers. In: IJCAI. pp. 5103--5111 (2018)

\bibitem{sill-nips97}
Sill, J.: Monotonic networks. In: NIPS. pp. 661--667 (1997)

\bibitem{vandenbroeck-nips20}
Sivaraman, A., Farnadi, G., Millstein, T.D., den Broeck, G.V.:
  Counterexample-guided learning of monotonic neural networks. In: NeurIPS
  (2020)

\bibitem{broeck-aaai2015}
{Van den Broeck}, G., Darwiche, A.: On the role of canonicity in knowledge
  compilation. In: AAAI. pp. 1641--1648 (2015)

\bibitem{kutyniok-jair21}
W{\"{a}}ldchen, S., MacDonald, J., Hauch, S., Kutyniok, G.: The computational
  complexity of understanding binary classifier decisions. J. Artif. Intell.
  Res.  \textbf{70},  351--387 (2021),
  \url{https://doi.org/10.1613/jair.1.12359}

\bibitem{wegener-bk00}
Wegener, I.: Branching Programs and Binary Decision Diagrams. {SIAM} (2000),
  \url{http://ls2-www.cs.uni-dortmund.de/monographs/bdd/}

\bibitem{louppe-nips19}
Wehenkel, A., Louppe, G.: Unconstrained monotonic neural networks. In: NeurIPS.
  pp. 1543--1553 (2019)

\bibitem{gupta-nips17}
You, S., Ding, D., Canini, K.R., Pfeifer, J., Gupta, M.R.: Deep lattice
  networks and partial monotonic functions. In: NeurIPS. pp. 2981--2989 (2017),
  \url{https://github.com/tensorflow/lattice}

\end{thebibliography}
}{
  % Import bibl (original .bbl) file

}
\label{lastpage}

\end{document}